\documentclass[10pt,twocolumn,letterpaper]{article}

\usepackage[pagenumbers]{cvpr}      %

\usepackage{cuted} %

\usepackage{currfile} %

\usepackage{caption} %
\usepackage{array} %
\usepackage{multirow} %

\makeatletter\patchcmd{\@maketitle}{\vspace*{12pt}}{}{}{\PackageError{arxiv}{title-block patch failed}{}}\makeatother   %

\definecolor{cvprblue}{rgb}{0.21,0.49,0.74}
\usepackage[pagebackref,breaklinks,colorlinks,allcolors=cvprblue]{hyperref}

\def\confName{CVPR}
\def\confYear{2027}
\def\arxiv{}   %

\title{Anchoring VAE Fine-Tuning to Its Latent Diffusion Prior}

\author{
Rafael Bischof$^{1}$ \quad Dominik Narnhofer$^{1}$ \quad Julius Erbach$^{1,2}$ \\
Michael Kraus$^{3}$ \quad Konrad Schindler$^{1}$ \quad Bernd Bickel$^{1}$ \\
$^{1}$ETH Zurich \quad $^{2}$MPI Informatics \quad $^{3}$TU Darmstadt
}

\begin{document}
\maketitle
\begin{strip}
\centering
\captionsetup{type=figure,hypcap=false}
\captionsetup[subfigure]{labelformat=empty}
\begin{subfigure}[b]{0.40202\linewidth}
  \includegraphics[width=\linewidth]{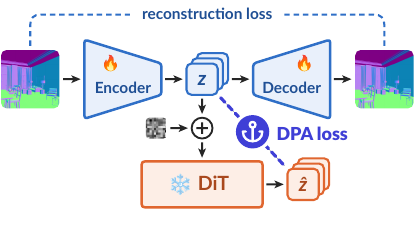}
  \caption{{\bf Step 1:} fine-tune the VAE with DPA}
  \label{fig:teaser_vae}
\end{subfigure}\hfill
\begin{subfigure}[b]{0.57778\linewidth}
  \includegraphics[width=\linewidth]{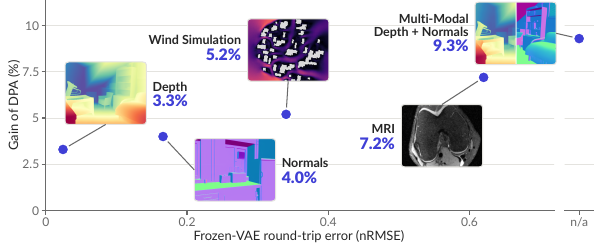}
  \caption{{\bf Step 2:} fine-tune the DiT on the new latents}
  \label{fig:teaser_dit}
\end{subfigure}
\caption{\textbf{Diffusion Prior Anchoring (DPA).}
\emph{Left:} During VAE adaptation, DPA pulls each latent~$z$ toward~$\hat z$, its denoised estimate generated by the frozen diffusion transformer (DiT).
\emph{Right:} Fine-tuning the DiT with the DPA latents yields gains that grow with the round-trip error of the frozen VAE. Multi-modal: gain on depth with a four-channel VAE, whose input the frozen VAE cannot encode (n/a).}
\captionsetup{hypcap=true}
\label{fig:\currfilebase}%
\end{strip}

\begin{abstract}
Pretrained latent diffusion models are effective priors for transfer learning, from geometry estimation to physics simulation.
Existing approaches fine-tune the diffusion backbone but freeze the encoder of the VAE to preserve the latent distribution.
However, a frozen encoder may miss salient details in non-RGB inputs and is limited to three channels.
We introduce \textbf{Diffusion Prior Anchoring (DPA)}, a regularizer applied during VAE fine-tuning that uses a score-distillation update to steer latents toward the high-density manifold on which the pretrained diffusion model denoises accurately.
We validate DPA across three image and video models (FLUX~3, LTX-Video, Wan2.2) and four tasks: monocular depth estimation, MRI reconstruction, wind-flow simulation, and surface normal estimation, where DPA lowers the prediction error by 7.5\% relative to the frozen VAE, 4.9\% to decoder-only fine-tuning, and 4.0\% to unanchored VAE fine-tuning.
Its gain grows the further the target data lies from the RGB training distribution, from 3.3\% on monocular depth estimation to 7.2\% on complex-valued MRI.
DPA also applies to modified VAEs, including pruned and multi-channel variants. With a four-channel VAE that encodes depth and surface normals, it lowers depth estimation error by 9.3\%.
\end{abstract}

\section{Introduction}
\label{sec:intro}

Latent diffusion models pretrained on large-scale image and video collections provide powerful generative priors for transfer learning to downstream tasks.
Fine-tuned on datasets far smaller than their pretraining corpora, they accurately estimate monocular depth and surface normals~\cite{ke2024marigold,he2025lotus,pavlovic2026marigoldv2}, simulate interactive environments in real time~\cite{valevski2025gamengen,he2025matrixgame2}, and serve as fast surrogates for physics simulations~\cite{perini2026windinet}.
In each case, transfer learning exploits the structural prior that the diffusion backbone has learned in the latent space of a separately trained VAE~\cite{rombach2022ldm}.

However, this paradigm rests on the assumption that the pretrained VAE accurately reconstructs the target domain.
In practice, this confines adaptation to three-channel inputs and outputs that resemble natural RGB images or videos.
In many domains of interest this assumption does not hold.
Geometric maps, complex-valued MRI scans, and physical velocity fields share the spatial and temporal layout of natural images and videos, but fall outside the RGB pretraining distribution.
Consequently, a frozen VAE reconstructs their pseudo-RGB encoding poorly. Modalities with more than three channels, such as joint depth and surface normals, cannot be encoded at all.

To unlock latent diffusion models for such domains one should adapt the VAE alongside the diffusion backbone.
In current practice only the VAE decoder is fine-tuned, while the encoder is kept frozen~\cite{valevski2025gamengen,pavlovic2026marigoldv2,perini2026windinet} to preserve the expected latent distribution for the pretrained diffusion model.
However, a decoder cannot recover information that the frozen encoder has not learned to preserve.
Jointly updating both the encoder and the decoder might seem the obvious solution, except that fine-tuning the VAE solely for reconstruction is unconstrained with respect to the diffusion model.
Latents thus drift away from the manifold on which the diffusion backbone was trained to denoise.

We propose \textbf{Diffusion Prior Anchoring (DPA)}, which regularizes VAE fine-tuning directly with the frozen diffusion model, via score distillation sampling (SDS)~\cite{poole2023dreamfusion}.
At each training step, an encoded latent is perturbed with noise, denoised by the frozen backbone, and pulled toward its clean estimate (\cref{fig:teaser}, left).
This steers the encoder toward latents that the pretrained prior recognizes and thus can denoise accurately, while the reconstruction loss preserves task-specific fidelity.
DPA needs only forward passes of the diffusion model and adds no inference cost.

We validate DPA across three image and video foundation models (FLUX~3, LTX-Video, Wan2.2) and four diverse tasks (\cref{sec:experiments}): surface normal estimation, monocular depth estimation, accelerated MRI reconstruction, and CFD urban wind-flow simulation.
DPA further extends to multi-modal prediction, where a four-channel VAE encodes depth and surface normals in a single shared latent, and to VAEs pruned by up to 80\% of their parameters.

Across all tasks and backbones, diffusion models trained on DPA-anchored latents outperform unanchored baselines, and the gains are largest where they are most needed, \ie, where the frozen VAE reconstructs the target domain poorly.
Systematic ablations show that anchoring is most beneficial when the diffusion backbone is fine-tuned with limited capacity, when multiple modalities share one latent, and when the backbone serves as an unconditional generative prior for posterior sampling.
Finally, we show that our setup is competitive with specialized state-of-the-art depth and surface normal estimators when trained with more data and pixel-space supervision.

In summary, our contributions are:
\begin{itemize}
    \item We introduce Diffusion Prior Anchoring (DPA), a backpropagation-free regularizer that aligns the VAE with the frozen diffusion prior, at zero inference cost.
    \item We systematically characterize when prior anchoring helps most and show that DPA improves monocular depth estimation, surface normal estimation, MRI reconstruction, and wind-flow simulation across three image and video backbones (FLUX~3, LTX-Video, Wan2.2).
    \item We show that DPA supports changes to the VAE architecture, such as changing the number of input and output channels and parameter pruning, while retaining the gain of anchoring.
\end{itemize}

\section{Related work}
\label{sec:related}

\paragraph{Repurposing latent diffusion models.}
Latent diffusion models for images~\cite{rombach2022ldm,esser2024sd3,wu2025qwenimage} and videos~\cite{blattmann2023stable,hacohen2024ltx,wan2025wan,wan22} learn generative priors over the compressed latent space of a pretrained VAE.
Fine-tuned on modest amounts of task-specific data, these foundation models become strong dense predictors for monocular depth and surface normals~\cite{ke2024marigold,fu2024geowizard,ye2024stablenormal,he2025lotus,garcia2025finetuning,pavlovic2026marigoldv2, ke2025video} as well as material properties~\cite{kocsis2024iid,zeng2024rgbx,ke2025marigold,liang2025diffusionrenderer}.
Beyond static scene understanding, pretrained image and video generators have been extended into world models that simulate complex environments for autonomous driving~\cite{gao2024vista} and robotics~\cite{nvidia2025cosmos,jang2025dreamgen}, game engines that respond to user actions in real time~\cite{valevski2025gamengen,li2025gamecraft,he2025matrixgame2}, and neural surrogates for physics simulators~\cite{perini2026windinet}.

\paragraph{VAE adaptation for new modalities.}
Existing approaches that repurpose latent diffusion models predominantly keep the pretrained VAE frozen~\cite{ke2024marigold,fu2024geowizard,he2025lotus,zeng2024rgbx,gao2024vista}.
This is natural for world models and game engines, which synthesize RGB frames within the VAE's pretraining distribution.
In contrast, dense predictors must cast non-RGB targets into three-channel pseudo-images compatible with the frozen encoder, \eg, by replicating single-channel depth maps across RGB channels or mapping surface normals to color space~\cite{ke2024marigold,fu2024geowizard,ye2024stablenormal,he2025lotus}. Multi-modal targets must be encoded into separate latents or predicted sequentially~\cite{fu2024geowizard,zeng2024rgbx,kocsis2024iid,ke2025marigold,liang2025diffusionrenderer}.
Even end-to-end pixel-space fine-tuning of the diffusion backbone has kept the VAE fixed~\cite{garcia2025finetuning}.
Where the frozen VAE reconstructs target modalities poorly, recent methods fine-tune the decoder, but keep the encoder frozen to preserve the latent distribution~\cite{valevski2025gamengen,pavlovic2026marigoldv2,perini2026windinet}, with Marigold~V2~\cite{pavlovic2026marigoldv2} further backpropagating decoded pixel-space losses through the updated decoder into the diffusion model.
In contrast, DPA enables channel expansion and pruning of both encoder and decoder, without disrupting the pretrained diffusion prior.

\paragraph{Latent space regularization and alignment.}
Several recent works regularize the VAE latent space so that a diffusion model fits it more easily.
REPA~\cite{yu2025repa} aligns the hidden states of a DiT with features of a pretrained vision foundation model, VA-VAE~\cite{yao2025vavae} applies this alignment to the VAE latents, and REPA-E~\cite{leng2025repae} uses it for joint end-to-end training of a VAE and a DiT from scratch.
Other regularizers enforce spatial~\cite{kouzelis2025eqvae} or scale equivariance~\cite{skorokhodov2025diffusability}.
These methods shape the latent space \emph{before} or \emph{during} the training of a new diffusion model.
Moreover, the alignment methods rely on vision encoders that are themselves trained on natural RGB images and thus face the same domain gap as the frozen VAE on non-RGB data.
In contrast, DPA adapts the VAE to an \emph{already pretrained} diffusion model and uses this frozen model itself as the regularizer.

\paragraph{Diffusion models as priors.}
Score distillation sampling (SDS)~\cite{poole2023dreamfusion} and Score Jacobian Chaining~\cite{wang2023sjc} optimize 3D representations by perturbing rendered views with noise and descending along the denoising residual of a frozen text-to-image diffusion model.
Subsequent variants refine the distillation target or noise schedule~\cite{wang2023prolificdreamer,hertz2023dds,huang2024dreamtime}, and related score-matching gradients have been used to train one-step generators~\cite{yin2024dmd} and personalized text-to-image adapters~\cite{arar2024palp}.
In inverse problems, diffusion posterior sampling~\cite{chung2023dps} and regularization by denoising (RED)~\cite{romano2017red,mardani2024reddiff,erbach2025flair} combine a measurement consistency term with a denoiser or diffusion model that acts as the prior.
DPA shares this structure, pairing the VAE reconstruction loss as a data fidelity term with the SDS gradient of a frozen diffusion prior, but applies it to regularize an encoder network over a dataset rather than optimizing individual reconstructions at test time.
While diffusion models have also been trained from scratch jointly with VAEs as learned latent priors~\cite{wehenkel2021diffusionpriors,vahdat2021lsgm,heek2026unifiedlatents}, DPA repurposes an existing pretrained diffusion transformer to regularize VAE adaptation with forward passes alone.

\section{Method}
\label{sec:method}

We start from a pretrained latent diffusion model~\cite{rombach2022ldm}, consisting of a VAE with encoder $E$ and decoder $D$, and a diffusion backbone $\phi$ that defines a prior $p_\phi$ over the latent space.
For an input $x$, we write $z=E(x)$ for the mean of the encoder's Gaussian posterior.
Given noisy latents $z_\lambda=\alpha_\lambda z+\sigma_\lambda\epsilon$ with $\epsilon\sim\mathcal{N}(0,I)$, indexed by the log signal-to-noise ratio (log-SNR) $\lambda=\log(\alpha_\lambda^2/\sigma_\lambda^2)$, let $\hat\epsilon_\phi(z_\lambda,\lambda)$ denote the diffusion model's noise prediction, obtained from its output by linear reparameterization.

\subsection{Diffusion Prior Anchoring}
\label{sec:anchor}
When the VAE is fine-tuned to a new domain, it is typically trained on $\mathcal{L}_\mathrm{VAE}$, which combines a domain-space reconstruction loss with KL regularization toward $\mathcal{N}(0,I)$.
Crucially, $\mathcal{L}_\mathrm{VAE}$ is agnostic to the pretrained diffusion model: besides the (typically weak) KL term, the encoder is free to map inputs to latent configurations that minimize reconstruction error, regardless of whether the frozen model $\phi$ can accurately denoise them.

Diffusion Prior Anchoring (DPA) connects VAE fine-tuning to the diffusion model by augmenting the training objective with an anchoring regularizer $\mathcal{A}$,
\begin{equation}
  \mathcal{L}_\mathrm{DPA}=\mathcal{L}_\mathrm{VAE}+\beta\,\mathbb{E}_x\big[\mathcal{A}\big(E(x)\big)\big],
  \label{eq:dpa}
\end{equation}
where $\beta>0$ controls the regularization strength.
Building on score distillation sampling (SDS)~\cite{poole2023dreamfusion}, we perturb $z$ into $z_\lambda$ and query the pretrained, frozen diffusion model $\phi$ for its clean-latent estimate $\hat z_\phi=(z_\lambda-\sigma_\lambda\hat\epsilon_\phi(z_\lambda,\lambda))/\alpha_\lambda$.
The anchor then regresses $z$ onto this estimate,
\begin{equation}
  \mathcal{A}(z)=\mathbb{E}_{\lambda,\epsilon}\Big[\tfrac{1}{2}\,w\big\|z-\mathrm{sg}[\hat z_\phi]\big\|^2\Big],
  \label{eq:anchor}
\end{equation}
where $w>0$ is a per-draw weight (\cref{sec:adjust}) and $\mathrm{sg}[\cdot]$ the stop-gradient operator, which treats $\hat z_\phi$ as a fixed target.
Differentiating \cref{eq:anchor} gives $\mathbb{E}_{\lambda,\epsilon}[w(z-\hat z_\phi)]$, and substituting $z_\lambda$ into $\hat z_\phi$ yields $z-\hat z_\phi=(\sigma_\lambda/\alpha_\lambda)(\hat\epsilon_\phi(z_\lambda,\lambda)-\epsilon)$.
Setting $w=\alpha_\lambda/\sigma_\lambda$ cancels this scale factor and recovers the unweighted SDS residual~\cite{poole2023dreamfusion},
\begin{equation}
  \nabla_z\mathcal{A}(z)=\mathbb{E}_{\lambda,\epsilon}\big[\hat\epsilon_\phi(z_\lambda,\lambda)-\epsilon\big].
  \label{eq:grad}
\end{equation}
As the diffusion model's prediction directly forms the gradient on $z$, each step requires only one forward pass through the frozen backbone and no backpropagation through $\phi$, with the expectation estimated from one draw of $(\lambda,\epsilon)$.

Perturbing $z$ into $z_\lambda$ maps adapted latents that have drifted from the pretraining manifold into regions of non-negligible density under the smoothed prior marginals $p_\lambda$~\cite{song2019ncsn}, where the frozen backbone remains accurate.
By Tweedie's formula~\cite{efron2011tweedie}, the optimal noise predictor equals $-\sigma_\lambda\nabla_{z_\lambda}\log p_\lambda(z_\lambda)$, which the backbone approximates via $\hat\epsilon_\phi(z_\lambda,\lambda)$~\cite{song2021sde}.
Since $\mathbb{E}[\epsilon]=0$, the gradient in \cref{eq:grad} is proportional to $-\nabla_z\mathbb{E}_\epsilon[\log p_\lambda(z_\lambda)]$ at each noise level, moving latents toward high-density regions of the smoothed prior~\cite{wang2023sjc} without evaluating $\log p_\phi(z)$, which is too expensive during training.
Although this mode-seeking update alone would collapse encodings toward the modes of $p_\phi$~\cite{wang2023prolificdreamer}, optimizing it jointly with $\mathcal{L}_\mathrm{VAE}$ preserves input-specific information for reconstruction while keeping latents in regions that $\phi$ can accurately denoise.

To anchor latents to the marginal prior $p_\phi$ rather than a specific condition, we query $\phi$ unconditionally.
For the image-editing and first-frame-conditioned video backbones in our experiments, this corresponds to passing an empty text prompt and omitting the reference image or first frame.

\subsection{Noise floor and sampling schedule}
\label{sec:adjust}
Instantiating the anchor in \cref{eq:anchor} requires addressing latent variations below the finest sampled noise level and specifying how $\lambda$ is sampled and weighted.

\paragraph{Noise floor.}
The anchor samples log-SNR levels up to $\lambda_\mathrm{max}$, corresponding to a minimum noise-to-signal ratio $\sigma_0=e^{-\lambda_\mathrm{max}/2}$.
Per-coordinate variations in $z$ much smaller than $\sigma_0$ are drowned out by noise at every sampled level and remain invisible to $\mathcal{A}$.
Without further constraint, the encoder can store high-frequency task details in sub-$\sigma_0$ variations, which lowers $\mathcal{L}_\mathrm{VAE}$ without increasing $\mathcal{A}$.
To prevent this, we impose a latent noise floor~\cite{heek2026unifiedlatents} during VAE training: the decoder reconstructs from $\tilde z+\sigma_0\epsilon_0$, where $\tilde z$ is a sample from the encoder's posterior and $\epsilon_0\sim\mathcal{N}(0,I)$, corrupting sub-$\sigma_0$ variations before decoding so they cannot reliably reduce reconstruction error.
The floor is applied only while the VAE is trained, so the DiT is trained on latents without the noise floor.

\paragraph{Noise sampling and weighting.}
Diffusion likelihood bounds motivate uniform coverage across log-SNR levels via $\lambda\sim\mathcal{U}[\lambda_\mathrm{min},\lambda_\mathrm{max}]$~\cite{kingma2023understanding,heek2026unifiedlatents}, whereas flow-matching diffusion transformers (DiTs) are pretrained under a non-uniform schedule $\pi(\lambda)$ concentrated at intermediate-to-high noise levels.
To combine broad scale coverage with the noise regime emphasized during DiT pretraining, we draw $\lambda$ from an equal mixture $\tfrac{1}{2}\,\mathcal{U}[\lambda_\mathrm{min},\lambda_\mathrm{max}]+\tfrac{1}{2}\,\pi$, with $\pi$ truncated to $[\lambda_\mathrm{min},\lambda_\mathrm{max}]$, which we empirically found to outperform either distribution alone (\cref{sec:exp_ablation}).
Draws from $\pi$ use $w=1/\sigma_\lambda$ instead of $\alpha_\lambda/\sigma_\lambda$, matching the flow-matching velocity residual (details in the supplement).

\section{Experiments}
\label{sec:experiments}

We first evaluate DPA on surface normal estimation against alternative VAE adaptation strategies (\cref{sec:exp_normals}), and use this setting to ablate the anchor (\cref{sec:exp_ablation}) and analyze how it changes the latent space (\cref{sec:exp_latent}).
We then modify the VAE architecture itself, expanding its input and output layers so that a single shared latent encodes both depth and surface normals (\cref{sec:exp_modify}) and pruning it for faster encoding and decoding (\cref{sec:exp_prune}), and compare DPA models trained at a larger scale with state-of-the-art methods (\cref{sec:exp_sota}).
Finally, we evaluate whether DPA's benefits transfer to non-RGB inverse problems and video foundation models through 12-fold accelerated MRI reconstruction (\cref{sec:exp_mri}) and CFD urban wind-flow simulation (\cref{sec:exp_wind}).

\subsection{Setup}
\label{sec:setup}
Every experiment follows a two-stage pipeline: we first adapt the VAE to the target domain and then fine-tune the DiT from its pretrained weights on the resulting latents.
With identical data, reconstruction losses, and step budgets across both stages, we compare four VAE variants:
(i)~\emph{Frozen}, the unmodified pretrained VAE,
(ii)~\emph{Decoder-only}, which fine-tunes $D$ with $E$ frozen,
(iii)~\emph{Control}, which fine-tunes $E$ and $D$ on $\mathcal{L}_\mathrm{VAE}$ without the anchor or noise floor, and
(iv)~\emph{DPA}, which fine-tunes $E$ and $D$ on $\mathcal{L}_\mathrm{DPA}$.
Round-trip reconstruction (\emph{Rec.}) evaluates $D(E(x))$ without the noise floor across all variants, and \emph{DiT loss} measures the flow-matching velocity loss of the frozen, unconditional pretrained DiT on the adapted latents prior to DiT fine-tuning.
Unless stated otherwise, DPA uses $[\lambda_\mathrm{min},\lambda_\mathrm{max}]=[-8,4]$ ($\sigma_0=e^{-2}\approx0.135$) and $\beta=1$, and each configuration is trained once.
Full training budgets and hyperparameters for all tasks are provided in the supplement.

\subsection{Surface normal estimation}
\label{sec:exp_normals}
For surface normal and monocular depth estimation, we fine-tune FLUX~3 Image Klein Base~\cite{bfl2026flux3} in image-editing mode with rank-128 LoRA on the flow-matching loss and sample with two Euler steps.
As in Marigold~V2~\cite{pavlovic2026marigoldv2}, the original, unmodified FLUX~3 VAE encodes the input RGB image into reference tokens, while the adapted VAE encodes and decodes the target map.
Both stages are trained on the synthetic indoor dataset Hypersim~\cite{roberts2021hypersim} and evaluated zero-shot via mean angular error on iBims-1~\cite{koch2018ibims}, NYUv2~\cite{silberman2012nyu}, and ScanNet~\cite{dai2017scannet}.

\begin{table}[t]
  \centering
  \small
  \caption{\textbf{Surface normal estimation.} Mean angular error (degrees). Rec.: round-trip reconstruction error of the VAE on Hypersim validation frames (degrees).}
  \label{tab:normals}
  \resizebox{\columnwidth}{!}{%
  \begin{tabular}{@{}lccccc@{}}
    \toprule
    VAE adaptation & Rec.$\downarrow$ & iBims-1$\downarrow$ & NYUv2$\downarrow$ & ScanNet$\downarrow$ & Mean$\downarrow$ \\
    \midrule
    Frozen & 3.54 & 15.93 & 17.53 & 15.12 & 16.19 \\
    Decoder-only & 2.16 & 15.41 & 16.95 & 14.90 & 15.75 \\
    Control & \textbf{1.54} & 15.33 & 16.87 & 14.64 & 15.61 \\
    DPA & 2.89 & \textbf{15.12} & \textbf{15.81} & \textbf{14.02} & \textbf{14.98} \\
    \bottomrule
  \end{tabular}%
  }
\end{table}

\begin{figure*}[t]
  \centering
  \includegraphics[width=\linewidth]{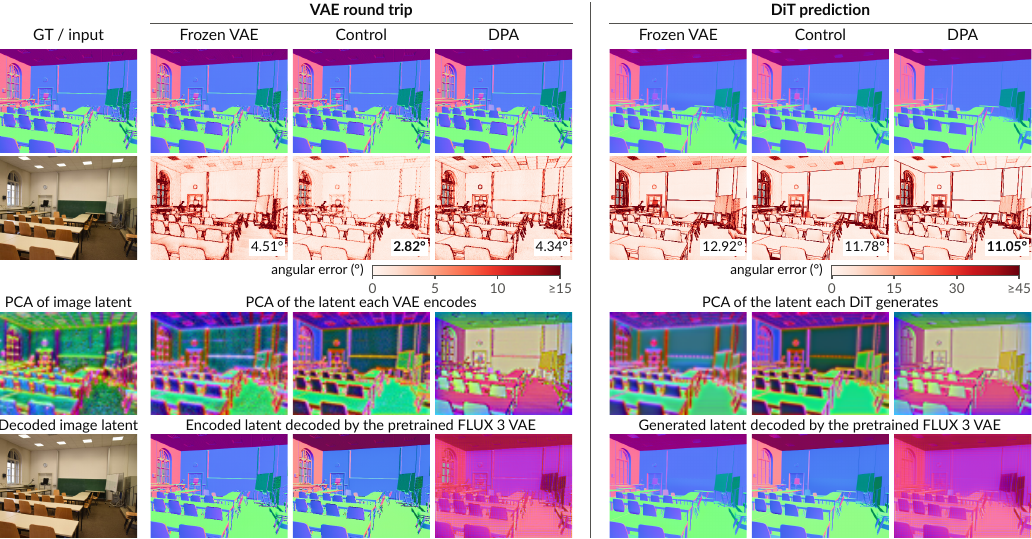}
  \caption{\textbf{Qualitative surface normals and latents} on an iBims-1 image. Round trip: the ground truth encoded and decoded by each VAE. Prediction: FLUX~3 fine-tuned on the latents of each VAE (\cref{tab:normals}). Second row: angular error maps with the mean error (degrees), the lowest per group in bold. Third row: first three principal components of the latents as RGB, with one basis and color range per VAE, shared by its encoded and generated latents. Fourth row: these latents decoded by the pretrained FLUX~3 VAE (the frozen VAE).}
  \label{fig:normals_qual}
\end{figure*}

\Cref{tab:normals} shows that, while unanchored fine-tuning cuts reconstruction error by more than half ($3.54^\circ$ to $1.54^\circ$), DPA trades some reconstruction fidelity ($2.89^\circ$) for prior alignment.
On the held-out benchmarks, however, the control lowers mean prediction error by only 3.6\% over the frozen VAE ($16.19^\circ$ to $15.61^\circ$), with three-quarters of that reduction already captured by adapting the decoder alone ($15.75^\circ$).
Anchoring the encoder with DPA brings mean error down to $14.98^\circ$ and achieves the lowest error on all three benchmarks, outperforming the control by $0.63^\circ$ (4.0\%) and more than doubling the total gain over the frozen baseline ($1.21^\circ$ vs.\ $0.58^\circ$).
\Cref{fig:normals_qual} shows an example, and the supplement shows images at fixed quantiles of the gain.

\subsection{Ablation studies}
\label{sec:exp_ablation}

As observed in \cref{tab:normals}, DPA improves downstream predictions despite incurring higher VAE reconstruction error.
To examine this trade-off, we vary the anchor weight $\beta$ from $0$ to $10$ in \cref{tab:weight} and report round-trip error, latent statistics, and the DiT loss, which quantifies how accurately the frozen DiT denoises adapted latents across noise levels.
Unanchored fine-tuning (control) increases the DiT loss by 40\% relative to the frozen VAE ($0.136$ to $0.190$), confirming that optimizing $\mathcal{L}_\mathrm{VAE}$ alone maps latents into regions that the pretrained backbone denoises less accurately.
Increasing $\beta$ monotonically reverses this effect, reducing the DiT loss to $0.074$ at $\beta=0.1$ and $0.029$ at $\beta=10$, while raising round-trip error from $1.54^\circ$ to $5.42^\circ$.
Downstream error is minimized between these extremes at $\beta=1$, where a more than fivefold reduction in DiT loss ($0.036$) outweighs the moderate increase in reconstruction error ($2.89^\circ$), though even weak regularization at $\beta=0.1$ improves upon the control by $0.27^\circ$.

With $\beta=1$ fixed, we ablate each design component of \cref{sec:method} individually in the bottom block of \cref{tab:weight} relative to our default configuration.
Every modification increases downstream error.
Conditioning the anchor on the input image rather than querying the unconditional prior causes the largest degradation ($+0.52^\circ$), followed by replacing the mixed schedule with uniform log-SNR sampling ($+0.35^\circ$), with both variants more than doubling the DiT loss ($0.090$ and $0.079$ vs.\ $0.036$).
Sampling exclusively from the pretraining schedule $\pi$ yields a smaller degradation ($+0.13^\circ$), whereas lowering the noise floor to $\sigma_0=0.05$ changes prediction error by only $0.05^\circ$ while reducing round-trip error from $2.89^\circ$ to $2.04^\circ$.

\begin{table}[t]
  \centering
  \small
  \caption{\textbf{DPA ablations.} Rec., Normals: round-trip and predicted mean angular error (degrees). Std, Dist.: standard deviation of the latents and their relative distance to those of the frozen encoder, on 16 validation frames. DiT loss: as in \cref{sec:setup}. Conditioned DPA: anchor conditioned on the input image.}
  \label{tab:weight}
  \resizebox{\columnwidth}{!}{%
  \begin{tabular}{@{}lccccc@{}}
    \toprule
    VAE adaptation & Rec.$\downarrow$ & Std & Dist. & DiT loss$\downarrow$ & Normals$\downarrow$ \\
    \midrule
    Frozen & 3.54 & 0.83 & 0.00 & 0.136 & 16.19 \\
    Control & \textbf{1.54} & 0.85 & 0.69 & 0.190 & 15.61 \\
    DPA, $\beta=0.1$ & 1.83 & 0.64 & 0.64 & 0.074 & 15.34 \\
    DPA, $\beta=1$ & 2.89 & 0.55 & 1.00 & 0.036 & \textbf{14.98} \\
    DPA, $\beta=10$ & 5.42 & 0.56 & 1.05 & \textbf{0.029} & 15.59 \\
    \midrule
    \multicolumn{6}{@{}l}{\emph{DPA, $\beta=1$, one change}} \\
    Uniform noise sampling & 2.22 & 0.76 & 0.58 & 0.079 & 15.34 \\
    $\pi$ noise sampling & 3.15 & 0.46 & 1.00 & 0.039 & 15.11 \\
    Noise floor: $\sigma_0=0.05$ & 2.04 & 0.55 & 1.01 & 0.034 & 15.03 \\
    Conditioned DPA & 2.34 & 0.68 & 0.64 & 0.090 & 15.50 \\
    \bottomrule
  \end{tabular}%
  }
\end{table}

\paragraph{DiT capacity.}
We next show in \cref{tab:capacity} that because DPA keeps latents aligned with the pretrained backbone, anchoring is most effective when the DiT's capacity to adapt is limited, such as with low-rank LoRA.
These runs fine-tune the DiT for half the steps of \cref{tab:normals}, so their errors are not directly comparable to it.
At rank 8, DPA reduces prediction error by $0.74^\circ$ (4.5\%) over the control, and as adaptation capacity increases through ranks 32 and 128 to full fine-tuning, this gain narrows monotonically to $0.36^\circ$.
Notably, DPA with rank-8 LoRA ($15.70^\circ$) already outperforms the unanchored control at every capacity level, including rank-128 LoRA ($15.85^\circ$) and full fine-tuning ($16.04^\circ$).

\begin{table}[t]
  \centering
  \small
  \caption{\textbf{DiT LoRA rank sweep.} Mean angular error (degrees) over iBims-1, NYUv2, and ScanNet, for LoRA of increasing rank and full fine-tuning.}
  \label{tab:capacity}
  \begin{tabular}{@{}lccc@{}}
    \toprule
    DiT fine-tuning & Control$\downarrow$ & DPA$\downarrow$ & DPA $-$ Control \\
    \midrule
    LoRA, rank 8 & 16.43 & \textbf{15.70} & $-0.74$ \\
    LoRA, rank 32 & 16.08 & \textbf{15.50} & $-0.58$ \\
    LoRA, rank 128 & 15.85 & \textbf{15.33} & $-0.52$ \\
    Full & 16.04 & \textbf{15.67} & $-0.36$ \\
    \bottomrule
  \end{tabular}
\end{table}

\begin{figure}[t]
  \centering
  \includegraphics[width=\linewidth]{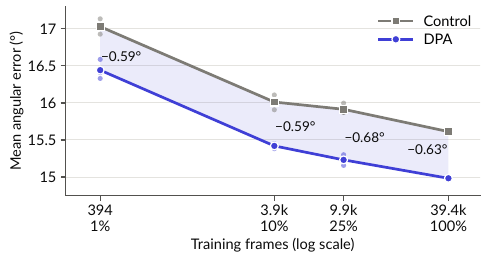}
  \caption{\textbf{Data efficiency.} Mean angular error (degrees) over iBims-1, NYUv2, and ScanNet against the number of Hypersim training frames. Dots: three random subsets per size. }
  \label{fig:data}
\end{figure}

\paragraph{Data efficiency.}
\label{sec:exp_data}
Turning from parameter budgets to sample efficiency, we evaluate in \cref{fig:data} how both stages scale when trained on nested random subsets of 1\%, 10\%, and 25\% of Hypersim across three random seeds.
Across two orders of magnitude in dataset size, from 394 frames (1\%) to all 39{,}415 frames, DPA reduces mean angular error relative to the control by a consistent $0.59^\circ$ to $0.68^\circ$, outperforming the unanchored baseline on every subset and seed.
At 10\% of Hypersim, DPA ($15.42^\circ$) already surpasses the control trained on 100\% of the dataset ($15.61^\circ$), providing a larger gain than a tenfold increase in training data.

\subsection{Latent space analysis}
\label{sec:exp_latent}

Besides lowering the DiT loss, DPA shrinks the latents (standard deviation $0.55$ against $0.85$ for the control) and moves them further from those of the frozen encoder (relative distance $1.00$ against $0.69$, \cref{tab:weight}).
Latents of smaller scale are easier to denoise, but scale explains only part of the lower DiT loss.
Rescaled to the standard deviation of the DPA latents, the control latents reach a DiT loss of $0.117$, three times the $0.036$ of DPA.
In the latents of \cref{fig:normals_qual} (third row), DPA encodes flat surfaces such as the walls and the floor as nearly uniform regions, whereas the latents of the frozen and control VAEs vary at fine scales there, and each DiT generates latents like those of its VAE.
The frozen VAE's decoder (fourth row) shows that the control latents remain close to the frozen VAE's encoding, which treats the normal map as an RGB image.
For a normal map this RGB encoding is arbitrary, and DPA changes it into one that suits the DiT better, with a DiT loss of $0.036$ instead of $0.190$ (\cref{tab:weight}) and a lower prediction error (\cref{tab:normals}).
To verify that DPA's gain stems from aligning latents with the diffusion prior rather than simpler confounding factors, we evaluate five modifications to the control VAE (latent rescaling, an L2 penalty toward the frozen latents, matched FLOPs, an angular loss, and VA-VAE alignment~\cite{yao2025vavae}). None improves upon the control (see supplement).

\subsection{VAE channel expansion for multi-modal data}
\label{sec:exp_modify}
\label{sec:exp_widen}
Many domain-specific tasks in physics, medical imaging, or 3D vision involve multi-modal data with more than three channels.
We study the effect of DPA on such tasks through joint monocular depth and surface normal estimation.
Initializing from the pretrained RGB kernels, we expand the first and last layers of the VAE to four channels and fine-tune it on concatenated depth and normal maps with and without DPA.
We then train a single DiT on this shared latent and compare it against separate single-modality models for depth and normals.

\begin{table}[t]
  \centering
  \small
  \renewcommand{\arraystretch}{1.1}   %
  \caption{\textbf{Multi-modal latent.} Channels: input channels of the VAE. Joint: one model on the latent of a four-channel VAE. Depth: zero-shot affine-invariant AbsRel ($\times100$), mean over NYUv2, KITTI~\cite{geiger2012kitti}, ETH3D~\cite{schops2017eth3d}, ScanNet, and DIODE~\cite{vasiljevic2019diode}, averaged over four sampling seeds. Normals: mean angular error (degrees) over iBims-1, NYUv2, and ScanNet.}
  \label{tab:widen}
  \begin{tabular}{@{}clccc@{}}
    \toprule
    & Method & Channels & Depth$\downarrow$ & Normals$\downarrow$ \\
    \midrule
    \multirow{2}{*}{\rotatebox[origin=c]{90}{\scriptsize\emph{Depth}}} & Control & 1 & 5.72 & \\
    & DPA & 1 & \textbf{5.52} & \\
    \midrule
    \multirow{2}{*}{\rotatebox[origin=c]{90}{\scriptsize\emph{Normals}}} & Control & 3 & & 15.49 \\
    & DPA & 3 & & 15.14 \\
    \midrule
    \multirow{2}{*}{\rotatebox[origin=c]{90}{\scriptsize\emph{Joint}}} & Control & 4 & 6.63 & 15.45 \\
    & DPA & 4 & 6.02 & \textbf{15.04} \\
    \bottomrule
  \end{tabular}
\end{table}

As shown in \cref{tab:widen}, DPA improves single-modality depth estimation, though by a smaller relative margin than on normals ($3.3\%$ vs.\ $4.0\%$ in \cref{tab:normals}), likely because the frozen VAE already reconstructs depth maps far better than normal maps (\cref{fig:teaser}).
The multi-modal model is less accurate on depth than the depth-only model, but slightly more accurate on normals than the normals-only model, for both the control and DPA.
Interestingly, the gain of DPA over the control is larger in the multi-modal model than with single modalities, $0.61$ vs.\ $0.19$ on depth and $0.40^\circ$ vs.\ $0.34^\circ$ on normals (differences are computed before rounding).
This again shows that DPA helps most where the data lies far from the training distribution of the pretrained VAE.
Tiling two latents of the frozen VAE, an alternative way of handling multiple modalities, matches the four-channel DPA model on depth ($6.01$ vs.\ $6.02$) but is $1.73^\circ$ less accurate on normals and needs twice as many latent tokens (see supplement).

\subsection{VAE pruning}
\label{sec:exp_prune}
Since the VAE is fine-tuned for the target domain, we can also experiment with pruning it.
Inspired by depth pruning of diffusion transformers~\cite{fang2025tinyfusion}, we remove whole residual and attention blocks from the FLUX~3 VAE, up to 80.1\% of its parameters, and fine-tune each pruned VAE on surface normals with and without DPA (details in the supplement).
Pruning raises the round-trip error but not the downstream error, and DPA stays $0.48^\circ$ to $0.65^\circ$ ahead of the control at every pruning level.
With 80.1\% of its parameters removed, the DPA VAE encodes and decodes $1.8\times$ faster and is still more accurate than the unpruned control ($15.00^\circ$ vs.\ $15.61^\circ$).

\subsection{Comparison with the state of the art}
\label{sec:exp_sota}
Although competing with specialized dense predictors is not the goal of this work, our setup is competitive with them when trained with more data and longer schedules (\cref{tab:final}).
We train depth, normals, and four-channel models on Hypersim and Virtual KITTI~2~\cite{cabon2020vkitti2}, followed by the second stage of Marigold~V2~\cite{pavlovic2026marigoldv2}, which fine-tunes the DiT and the VAE decoder with pixel-space losses for up to 3{,}000 steps.
Our normal estimates are the most accurate in both blocks of \cref{tab:final}, and our depth estimates are second only to Marigold~V2 and FE2E~\cite{wang2026fe2e}, respectively.
Per-benchmark results and training details are in the supplement.

\begin{table}[t]
  \centering
  \small
  \caption{\textbf{Comparison with SOTA.} Metrics as in \cref{tab:widen}, with Sintel~\cite{butler2012sintel} added for normals. Images: training frames. Baselines' numbers taken from~\cite{pavlovic2026marigoldv2}. Ours averaged over four seeds.}
  \label{tab:final}
  \begin{tabular}{@{}clccc@{}}
    \toprule
    & Method & Images & Depth$\downarrow$ & Normals$\downarrow$ \\
    \midrule
    \multirow{3}{*}{\rotatebox[origin=c]{90}{\emph{Depth}}} & Marigold~V2~\cite{pavlovic2026marigoldv2} & 74K & \textbf{4.14} & \\
    & Lotus-2~\cite{he2025lotus2} & 59K & 5.00 & \\
    & FLUX~3 + DPA (ours) & 90K & 4.39 & \\
    \midrule
    \multirow{3}{*}{\rotatebox[origin=c]{90}{\emph{Normals}}} & Marigold~V2~\cite{pavlovic2026marigoldv2} & 59K & & 18.83 \\
    & Lotus-2~\cite{he2025lotus2} & 59K & & 19.20 \\
    & FLUX~3 + DPA (ours) & 61K & & \textbf{17.84} \\
    \midrule
    \multirow{3}{*}{\rotatebox[origin=c]{90}{\emph{Joint}}} & FE2E~\cite{wang2026fe2e} & 71K & \textbf{4.80} & 19.08 \\
    & GeoWizard~\cite{fu2024geowizard} & 280K & 7.88 & 23.98 \\
    & FLUX~3 + DPA (ours) & 61K & 4.89 & \textbf{17.61} \\
    \bottomrule
  \end{tabular}
\end{table}

\subsection{Accelerated MRI reconstruction}
\label{sec:exp_mri}
Beyond 3D scene geometry, DPA applies to non-RGB modalities in domains such as medical imaging.
We evaluate it on accelerated MRI, which reduces the time that patients spend inside the machine by acquiring only a subset of the Fourier-domain ($k$-space) measurements and reconstructing the image from this undersampled data.
Here, the DiT serves as an unconditional generative prior rather than a conditional predictor: it is trained solely on clean slices, while undersampled measurements enter only during posterior sampling at inference~\cite{chung2023dps}.
Following~\cite{desai2023noise2recon}, multi-coil knee volumes from mridata.org~\cite{epperson2013mridata} are split into 14 training, 2 validation, and 3 test scans at 12-fold undersampling.
We encode each complex 2D slice as real, imaginary, and zero channels in the FLUX~3 VAE, set $\beta=0.1$ for DPA, and fine-tune the DiT unconditionally.
Inference alternates latent denoising with $k$-space data consistency on the decoded complex image via decomposed diffusion sampling~\cite{chung2024dds,song2024resample}, evaluated with PSNR and SSIM across all 96 test slices.

\begin{table}[t]
  \centering
  \small
  \caption{\textbf{Accelerated MRI reconstruction} at 12-fold undersampling. Magnitude PSNR (dB) and SSIM. Rec.: round-trip PSNR of the VAE. Bottom block: FLUX~3 fine-tuned on the latents of each VAE.}
  \label{tab:mri}
  \begin{tabular}{@{}lccc@{}}
    \toprule
    Method & Rec.$\uparrow$ & PSNR$\uparrow$ & SSIM$\uparrow$ \\
    \midrule
    Zero-filled & -- & 26.04 & 0.666 \\
    L1-wavelet~\cite{lustig2007sparse} & -- & 29.18 & 0.720 \\
    Langevin~\cite{jalal2021robust} & -- & 25.80 & 0.659 \\
    \midrule
    Frozen VAE & 28.55 & 28.29 & 0.704 \\
    Decoder-only & 29.76 & 28.78 & 0.711 \\
    Control & \textbf{31.39} & 28.87 & 0.700 \\
    DPA & 28.98 & \textbf{29.52} & \textbf{0.736} \\
    \bottomrule
  \end{tabular}
\end{table}

\begin{figure*}[t]
  \centering
  \includegraphics[width=\linewidth]{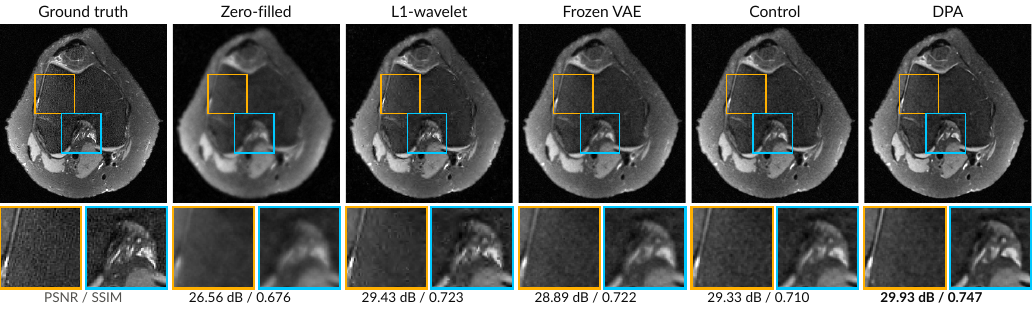}
  \caption{\textbf{Knee MRI reconstruction at 12-fold acceleration.} Test slice with the median PSNR gain of DPA over the control, with two zoomed regions. Methods as in \cref{tab:mri}. Bottom: magnitude PSNR (dB) / SSIM, best in bold.}
  \label{fig:mri}
\end{figure*}

In \cref{tab:mri}, vanilla FLUX~3 trails classical L1-wavelet compressed sensing by $0.89$\,dB with the frozen VAE and by $0.31$\,dB with the unanchored control.
Anchoring the VAE with DPA surpasses L1-wavelet by $0.34$\,dB and outperforms the control by $0.65$\,dB, improving PSNR on all 96 test slices and recovering sharper anatomical structure (\cref{fig:mri}).
Posterior reconstructions can exceed the VAE's round-trip PSNR because the final data-consistency step restores the measured $k$-space frequencies.
The supplement reports several ablations, sampler details, and hyperparameter sweeps.

\subsection{Urban wind-flow simulation}
\label{sec:exp_wind}
Finally, we test whether DPA extends to video and to backbones other than FLUX~3, with two video diffusion models: LTX-Video~\cite{hacohen2024ltx} and the $2.5\times$ larger Wan2.2-TI2V-5B~\cite{wan22}.
Both are evaluated on the urban wind-flow task of WinDiNet~\cite{perini2026windinet}, where a model conditioned on a 2D building footprint, inlet wind speed, and domain scale predicts wind velocities over 112 time steps.
Whereas WinDiNet fine-tunes only the decoder of LTX-Video, we jointly adapt the encoder and decoder of both models, setting $\beta=10^{-2}$ for DPA before fine-tuning each DiT.
Rollout accuracy is evaluated via the variance-normalized RMSE (VRMSE)~\cite{perini2026windinet} on WinDiNet's test set.
\begin{table}[t]
  \centering
  \small
  \caption{\textbf{Urban wind-flow simulation.} Rollout VRMSE over 1{,}000 test CFD simulations. Rec.: round-trip VRMSE of the video VAE on test simulations. WinDiNet: official released checkpoint, not trained under our recipe.}
  \label{tab:wind}
  \resizebox{\columnwidth}{!}{%
  \begin{tabular}{@{}llcc@{}}
    \toprule
    Backbone & VAE adaptation & Rec.$\downarrow$ & Test$\downarrow$ \\
    \midrule
    LTX-Video & Decoder-only (WinDiNet)~\cite{perini2026windinet} & \textbf{0.141} & 0.521 \\
    LTX-Video & Control & 0.158 & 0.478 \\
    LTX-Video & DPA & 0.164 & \textbf{0.453} \\
    \midrule
    Wan2.2 & Control & \textbf{0.148} & 0.450 \\
    Wan2.2 & DPA & 0.165 & \textbf{0.428} \\
    \bottomrule
  \end{tabular}%
  }
\end{table}

\begin{figure}[t]
  \centering
  \includegraphics[width=\linewidth]{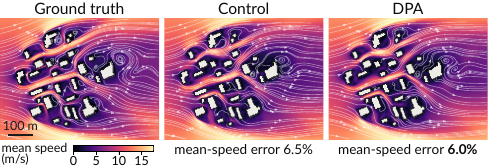}
  \caption{\textbf{Wan2.2 wind-flow predictions} on the test simulation with the median gain of DPA. Color: time-averaged wind speed. Lines: streamlines of the time-averaged velocity. Error: relative error of the time-averaged speed.}
  \label{fig:wind}
\end{figure}

\Cref{tab:wind} shows that DPA also benefits this CFD task, reducing the test VRMSE of the control by 5.2\% with LTX-Video and by 4.9\% with Wan2.2 while lowering the error on 75\% and 81\% of the test simulations, respectively.
Wan2.2 with DPA performs best, 17.9\% below the released WinDiNet checkpoint.
\Cref{fig:wind} shows time-averaged flow predicted by both Wan2.2 models.
The supplement details the setup and shows that, on average, building layouts optimized with the DPA surrogate improve the simulated flow the most.

\subsection{Training cost}
\label{sec:exp_cost}
The anchor adds one forward pass of the frozen DiT per VAE training step, which increases the VAE training time by 39\% for FLUX~3 and by 4\% for LTX-Video, where the video VAE step dominates the cost. To test whether this additional compute is responsible for the gain of DPA, we train a FLOPs-matched control on surface normals ($1.9\times$ the VAE steps). It lowers the round-trip error from $1.54^\circ$ to $1.42^\circ$ but not the prediction error ($15.71^\circ$ vs.\ $14.98^\circ$ for DPA), so the gain stems from the anchor (see supplement).

\section{Conclusion}
\label{sec:conclusion}
Adapting a latent diffusion model to a new modality usually keeps the VAE encoder frozen to preserve compatibility with the pretraining prior.
On out-of-distribution inputs, however, a frozen encoder loses salient detail and restricts adaptation to modalities that can be mapped onto three RGB channels.
Diffusion Prior Anchoring (DPA) removes this bottleneck by using the frozen diffusion model itself to regularize VAE training via score distillation.
It needs only forward passes of the frozen DiT, adds no cost at inference, and supports adapted, channel-expanded, and pruned VAEs across image and video backbones.

DPA improved downstream accuracy across every modality we tested, with gains that correlate with the reconstruction error of the frozen VAE (\cref{fig:teaser}).
Where the pretrained VAE already reconstructs inputs well, as in monocular depth estimation, the improvement is modest at 3.3\%, whereas domains with large round-trip error see substantially larger gains, reaching 7.2\% on MRI.
Within a given task, anchoring helps most when the model relies strongly on the pretrained prior, such as under low-rank fine-tuning or unconditional sampling, while its benefit narrows when full fine-tuning has enough capacity to absorb latent drift.

Our results also raise open questions for future work.
First, although SDS is known to be mode-seeking, replacing it in DPA with objectives that address this limitation~\cite{wang2023prolificdreamer,erbach2025flair} did not improve downstream accuracy in our supplementary experiments.
Combining SDS with the VAE reconstruction loss may already prevent mode collapse, yet a more systematic study of score-distillation variants in this setting could still yield stronger anchors.
Second, the unconditional prior proved a better anchor than one conditioned on the input image, even though conditioning matches the downstream setup more closely.
Understanding this behavior, together with anchoring to priors that already model the target task, is a promising direction for future research.

\section*{Acknowledgements}
This work was supported as part of the Swiss AI Initiative by small grants from the Swiss National Supercomputing Centre (CSCS) under project IDs a0187 and a0250 on Alps.
Julius Erbach was supported by the Max Planck ETH Center for Learning Systems (CLS).

\section*{LLM Usage Statement}
We used the Claude Code assistant (Anthropic) in three supporting roles.
It wrote plotting and figure-layout code.
It implemented and debugged parts of the code base and ran some of the experiments to the authors' specifications.
It suggested edits to individual sentences and captions, which the authors revised.
The authors checked all code, results, and text, and take full responsibility for the paper.

{
    \small
    \bibliographystyle{ieeenat_fullname}
    \bibliography{main}

\begin{thebibliography}{76}
\providecommand{\natexlab}[1]{#1}
\providecommand{\url}[1]{\texttt{#1}}
\expandafter\ifx\csname urlstyle\endcsname\relax
  \providecommand{\doi}[1]{doi: #1}\else
  \providecommand{\doi}{doi: \begingroup \urlstyle{rm}\Url}\fi

\bibitem[Arar et~al.(2024)Arar, Voynov, Hertz, Avrahami, Fruchter, Pritch,
  Cohen-Or, and Shamir]{arar2024palp}
Moab Arar, Andrey Voynov, Amir Hertz, Omri Avrahami, Shlomi Fruchter, Yael
  Pritch, Daniel Cohen-Or, and Ariel Shamir.
\newblock {PALP}: Prompt aligned personalization of text-to-image models.
\newblock In \emph{SIGGRAPH Asia Conference Papers}, 2024.

\bibitem[Bae and Davison(2024)]{bae2024dsine}
Gwangbin Bae and Andrew~J. Davison.
\newblock Rethinking inductive biases for surface normal estimation.
\newblock In \emph{IEEE/CVF Conference on Computer Vision and Pattern
  Recognition (CVPR)}, 2024.

\bibitem[{Black Forest Labs}(2026)]{bfl2026flux3}
{Black Forest Labs}.
\newblock {FLUX 3 Image Klein}.
\newblock \url{https://bfl.ai/blog/flux-3-image-klein}, 2026.

\bibitem[Blattmann et~al.(2023)Blattmann, Dockhorn, Kulal, Mendelevitch,
  Kilian, Lorenz, Levi, English, Voleti, Letts, Jampani, and
  Rombach]{blattmann2023stable}
Andreas Blattmann, Tim Dockhorn, Sumith Kulal, Daniel Mendelevitch, Maciej
  Kilian, Dominik Lorenz, Yam Levi, Zion English, Vikram Voleti, Adam Letts,
  Varun Jampani, and Robin Rombach.
\newblock {Stable} {Video} {Diffusion}: Scaling latent video diffusion models
  to large datasets.
\newblock \emph{preprint arXiv:2311.15127}, 2023.

\bibitem[Butler et~al.(2012)Butler, Wulff, Stanley, and
  Black]{butler2012sintel}
Daniel~J. Butler, Jonas Wulff, Garrett~B. Stanley, and Michael~J. Black.
\newblock A naturalistic open source movie for optical flow evaluation.
\newblock In \emph{European Conference on Computer Vision (ECCV)}, 2012.

\bibitem[Cabon et~al.(2020)Cabon, Murray, and Humenberger]{cabon2020vkitti2}
Yohann Cabon, Naila Murray, and Martin Humenberger.
\newblock Virtual {KITTI} 2.
\newblock \emph{preprint arXiv:2001.10773}, 2020.

\bibitem[Chen et~al.(2018)Chen, Badrinarayanan, Lee, and
  Rabinovich]{chen2018gradnorm}
Zhao Chen, Vijay Badrinarayanan, Chen-Yu Lee, and Andrew Rabinovich.
\newblock {GradNorm}: Gradient normalization for adaptive loss balancing in
  deep multitask networks.
\newblock In \emph{International Conference on Machine Learning (ICML)}, 2018.

\bibitem[Chung et~al.(2023)Chung, Kim, McCann, Klasky, and Ye]{chung2023dps}
Hyungjin Chung, Jeongsol Kim, Michael~T. McCann, Marc~L. Klasky, and Jong~Chul
  Ye.
\newblock Diffusion posterior sampling for general noisy inverse problems.
\newblock In \emph{International Conference on Learning Representations
  (ICLR)}, 2023.

\bibitem[Chung et~al.(2024)Chung, Lee, and Ye]{chung2024dds}
Hyungjin Chung, Suhyeon Lee, and Jong~Chul Ye.
\newblock Decomposed diffusion sampler for accelerating large-scale inverse
  problems.
\newblock In \emph{International Conference on Learning Representations
  (ICLR)}, 2024.

\bibitem[Dai et~al.(2017)Dai, Chang, Savva, Halber, Funkhouser, and
  Nie{\ss}ner]{dai2017scannet}
Angela Dai, Angel~X. Chang, Manolis Savva, Maciej Halber, Thomas Funkhouser,
  and Matthias Nie{\ss}ner.
\newblock {ScanNet}: Richly-annotated {3D} reconstructions of indoor scenes.
\newblock In \emph{IEEE Conference on Computer Vision and Pattern Recognition
  (CVPR)}, 2017.

\bibitem[Desai et~al.(2023)Desai, Ozturkler, Sandino, Boutin, Willis,
  Vasanawala, Hargreaves, R{\'e}, Pauly, and Chaudhari]{desai2023noise2recon}
Arjun~D. Desai, Batu~M. Ozturkler, Christopher~M. Sandino, Robert Boutin, Marc
  Willis, Shreyas Vasanawala, Brian~A. Hargreaves, Christopher R{\'e}, John~M.
  Pauly, and Akshay~S. Chaudhari.
\newblock {Noise2Recon}: Enabling {SNR}-robust {MRI} reconstruction with
  semi-supervised and self-supervised learning.
\newblock \emph{Magnetic Resonance in Medicine}, 90\penalty0 (5):\penalty0
  2052--2070, 2023.

\bibitem[Efron(2011)]{efron2011tweedie}
Bradley Efron.
\newblock Tweedie's formula and selection bias.
\newblock \emph{Journal of the American Statistical Association}, 106\penalty0
  (496):\penalty0 1602--1614, 2011.

\bibitem[Epperson et~al.(2013)Epperson, Sawyer, Lustig, Alley, Uecker, Virtue,
  Lai, and Vasanawala]{epperson2013mridata}
Kevin Epperson, Anne~Marie Sawyer, Michael Lustig, Marcus~T. Alley, Martin
  Uecker, Patrick Virtue, Peng Lai, and Shreyas~S. Vasanawala.
\newblock Creation of fully sampled {MR} data repository for compressed sensing
  of the knee.
\newblock In \emph{SMRT Annual Meeting}, 2013.

\bibitem[Erbach et~al.(2025)Erbach, Narnhofer, Dombos, Schiele, Lenssen, and
  Schindler]{erbach2025flair}
Julius Erbach, Dominik Narnhofer, Andreas Dombos, Bernt Schiele, Jan~Eric
  Lenssen, and Konrad Schindler.
\newblock Solving inverse problems with {FLAIR}.
\newblock In \emph{Advances in Neural Information Processing Systems
  (NeurIPS)}, 2025.

\bibitem[Esser et~al.(2024)Esser, Kulal, Blattmann, Entezari,
  et~al.]{esser2024sd3}
Patrick Esser, Sumith Kulal, Andreas Blattmann, Rahim Entezari, et~al.
\newblock Scaling rectified flow transformers for high-resolution image
  synthesis.
\newblock In \emph{International Conference on Machine Learning (ICML)}, 2024.

\bibitem[Fang et~al.(2025)Fang, Li, Ma, and Wang]{fang2025tinyfusion}
Gongfan Fang, Kunjun Li, Xinyin Ma, and Xinchao Wang.
\newblock {TinyFusion}: Diffusion transformers learned shallow.
\newblock In \emph{IEEE/CVF Conference on Computer Vision and Pattern
  Recognition (CVPR)}, 2025.

\bibitem[Fu et~al.(2024)Fu, Yin, Hu, Wang, Ma, Tan, Shen, Lin, and
  Long]{fu2024geowizard}
Xiao Fu, Wei Yin, Mu Hu, Kaixuan Wang, Yuexin Ma, Ping Tan, Shaojie Shen, Dahua
  Lin, and Xiaoxiao Long.
\newblock {GeoWizard}: Unleashing the diffusion priors for {3D} geometry
  estimation from a single image.
\newblock In \emph{European Conference on Computer Vision (ECCV)}, 2024.

\bibitem[Gao et~al.(2024)Gao, Yang, Chen, Chitta, Qiu, Geiger, Zhang, and
  Li]{gao2024vista}
Shenyuan Gao, Jiazhi Yang, Li Chen, Kashyap Chitta, Yihang Qiu, Andreas Geiger,
  Jun Zhang, and Hongyang Li.
\newblock Vista: A generalizable driving world model with high fidelity and
  versatile controllability.
\newblock In \emph{Advances in Neural Information Processing Systems
  (NeurIPS)}, 2024.

\bibitem[Geiger et~al.(2012)Geiger, Lenz, and Urtasun]{geiger2012kitti}
Andreas Geiger, Philip Lenz, and Raquel Urtasun.
\newblock Are we ready for autonomous driving? {T}he {KITTI} vision benchmark
  suite.
\newblock In \emph{IEEE Conference on Computer Vision and Pattern Recognition
  (CVPR)}, 2012.

\bibitem[HaCohen et~al.(2024)HaCohen, Chiprut, Brazowski, Shalem, Moshe,
  Richardson, Levin, Shiran, Zabari, Gordon, et~al.]{hacohen2024ltx}
Yoav HaCohen, Nisan Chiprut, Benny Brazowski, Daniel Shalem, Dudu Moshe, Eitan
  Richardson, Eran Levin, Guy Shiran, Nir Zabari, Ori Gordon, et~al.
\newblock {LTX-Video}: Realtime video latent diffusion.
\newblock \emph{preprint arXiv:2501.00103}, 2024.

\bibitem[He et~al.(2025{\natexlab{a}})He, Li, Sheng, and Chen]{he2025lotus2}
Jing He, Haodong Li, Mingzhi Sheng, and Ying-Cong Chen.
\newblock Lotus-2: Advancing geometric dense prediction with powerful image
  generative model.
\newblock \emph{preprint arXiv:2512.01030}, 2025{\natexlab{a}}.

\bibitem[He et~al.(2025{\natexlab{b}})He, Li, Yin, Liang, Li, Zhou, Zhang, Liu,
  and Chen]{he2025lotus}
Jing He, Haodong Li, Wei Yin, Yixun Liang, Leheng Li, Kaiqiang Zhou, Hongbo
  Zhang, Bingbing Liu, and Ying-Cong Chen.
\newblock Lotus: Diffusion-based visual foundation model for high-quality dense
  prediction.
\newblock In \emph{International Conference on Learning Representations
  (ICLR)}, 2025{\natexlab{b}}.

\bibitem[He et~al.(2025{\natexlab{c}})He, Peng, Liu, Wang, Zhang, Cui, Kang,
  Jiang, An, Ren, Xu, Guo, Gong, Wu, Li, Song, Liu, Li, and
  Zhou]{he2025matrixgame2}
Xianglong He, Chunli Peng, Zexiang Liu, Boyang Wang, Yifan Zhang, Qi Cui, Fei
  Kang, Biao Jiang, Mengyin An, Yangyang Ren, Baixin Xu, Hao-Xiang Guo,
  Kaixiong Gong, Size Wu, Wei Li, Xuchen Song, Yang Liu, Yangguang Li, and
  Yahui Zhou.
\newblock {Matrix-Game 2.0}: An open-source, real-time, and streaming
  interactive world model.
\newblock \emph{preprint arXiv:2508.13009}, 2025{\natexlab{c}}.

\bibitem[Heek et~al.(2026)Heek, Hoogeboom, Mensink, and
  Salimans]{heek2026unifiedlatents}
Jonathan Heek, Emiel Hoogeboom, Thomas Mensink, and Tim Salimans.
\newblock {Unified Latents} ({UL}): How to train your latents.
\newblock \emph{preprint arXiv:2602.17270}, 2026.

\bibitem[Hertz et~al.(2023)Hertz, Aberman, and Cohen-Or]{hertz2023dds}
Amir Hertz, Kfir Aberman, and Daniel Cohen-Or.
\newblock Delta denoising score.
\newblock In \emph{IEEE/CVF International Conference on Computer Vision
  (ICCV)}, 2023.

\bibitem[Huang et~al.(2024)Huang, Wang, Shi, Tang, Qi, and
  Zhang]{huang2024dreamtime}
Yukun Huang, Jianan Wang, Yukai Shi, Boshi Tang, Xianbiao Qi, and Lei Zhang.
\newblock {DreamTime}: An improved optimization strategy for diffusion-guided
  {3D} generation.
\newblock In \emph{International Conference on Learning Representations
  (ICLR)}, 2024.

\bibitem[Jalal et~al.(2021)Jalal, Arvinte, Daras, Price, Dimakis, and
  Tamir]{jalal2021robust}
Ajil Jalal, Marius Arvinte, Giannis Daras, Eric Price, Alexandros~G. Dimakis,
  and Jonathan~I. Tamir.
\newblock Robust compressed sensing {MRI} with deep generative priors.
\newblock In \emph{Advances in Neural Information Processing Systems
  (NeurIPS)}, 2021.

\bibitem[Jang et~al.(2025)Jang, Ye, Lin, Xiang, Bjorck, Fang, Hu, Huang,
  Kundalia, Lin, Magne, Mandlekar, Narayan, Tan, Wang, Wang, Wang, Xu, Zeng,
  Zheng, Zheng, Liu, Zettlemoyer, Fox, Kautz, Reed, Zhu, and
  Fan]{jang2025dreamgen}
Joel Jang, Seonghyeon Ye, Zongyu Lin, Jiannan Xiang, Johan Bjorck, Yu Fang,
  Fengyuan Hu, Spencer Huang, Kaushil Kundalia, Yen-Chen Lin, Lo{\"\i}c Magne,
  Ajay Mandlekar, Avnish Narayan, You~Liang Tan, Guanzhi Wang, Jing Wang, Qi
  Wang, Yinzhen Xu, Xiaohui Zeng, Kaiyuan Zheng, Ruijie Zheng, Ming-Yu Liu,
  Luke Zettlemoyer, Dieter Fox, Jan Kautz, Scott Reed, Yuke Zhu, and Linxi Fan.
\newblock {DreamGen}: Unlocking generalization in robot learning through video
  world models.
\newblock In \emph{Conference on Robot Learning (CoRL)}, 2025.

\bibitem[Ke et~al.(2024)Ke, Obukhov, Huang, Metzger, Daudt, and
  Schindler]{ke2024marigold}
Bingxin Ke, Anton Obukhov, Shengyu Huang, Nando Metzger, Rodrigo~Caye Daudt,
  and Konrad Schindler.
\newblock Repurposing diffusion-based image generators for monocular depth
  estimation.
\newblock In \emph{IEEE/CVF Conference on Computer Vision and Pattern
  Recognition (CVPR)}, 2024.

\bibitem[Ke et~al.(2025{\natexlab{a}})Ke, Narnhofer, Huang, Ke, Peters,
  Fragkiadaki, Obukhov, and Schindler]{ke2025video}
Bingxin Ke, Dominik Narnhofer, Shengyu Huang, Lei Ke, Torben Peters, Katerina
  Fragkiadaki, Anton Obukhov, and Konrad Schindler.
\newblock Video depth without video models.
\newblock In \emph{IEEE/CVF Conference on Computer Vision and Pattern
  Recognition (CVPR)}, 2025{\natexlab{a}}.

\bibitem[Ke et~al.(2025{\natexlab{b}})Ke, Qu, Wang, Metzger, Huang, Li,
  Obukhov, and Schindler]{ke2025marigold}
Bingxin Ke, Kevin Qu, Tianfu Wang, Nando Metzger, Shengyu Huang, Bo Li, Anton
  Obukhov, and Konrad Schindler.
\newblock Marigold: Affordable adaptation of diffusion-based image generators
  for image analysis.
\newblock \emph{IEEE Transactions on Pattern Analysis and Machine Intelligence
  (TPAMI)}, 2025{\natexlab{b}}.

\bibitem[Kingma and Gao(2023)]{kingma2023understanding}
Diederik~P. Kingma and Ruiqi Gao.
\newblock Understanding diffusion objectives as the {ELBO} with simple data
  augmentation.
\newblock In \emph{Advances in Neural Information Processing Systems
  (NeurIPS)}, 2023.

\bibitem[Kingma et~al.(2021)Kingma, Salimans, Poole, and Ho]{kingma2021vdm}
Diederik~P. Kingma, Tim Salimans, Ben Poole, and Jonathan Ho.
\newblock Variational diffusion models.
\newblock In \emph{Advances in Neural Information Processing Systems
  (NeurIPS)}, 2021.

\bibitem[Koch et~al.(2018)Koch, Liebel, Fraundorfer, and
  K{\"o}rner]{koch2018ibims}
Tobias Koch, Lukas Liebel, Friedrich Fraundorfer, and Marco K{\"o}rner.
\newblock Evaluation of {CNN}-based single-image depth estimation methods.
\newblock In \emph{European Conference on Computer Vision Workshops (ECCVW)},
  2018.

\bibitem[Kocsis et~al.(2024)Kocsis, Sitzmann, and Nie{\ss}ner]{kocsis2024iid}
Peter Kocsis, Vincent Sitzmann, and Matthias Nie{\ss}ner.
\newblock Intrinsic image diffusion for indoor single-view material estimation.
\newblock In \emph{IEEE/CVF Conference on Computer Vision and Pattern
  Recognition (CVPR)}, 2024.

\bibitem[Kouzelis et~al.(2025)Kouzelis, Kakogeorgiou, Gidaris, and
  Komodakis]{kouzelis2025eqvae}
Theodoros Kouzelis, Ioannis Kakogeorgiou, Spyros Gidaris, and Nikos Komodakis.
\newblock {EQ-VAE}: Equivariance regularized latent space for improved
  generative image modeling.
\newblock In \emph{International Conference on Machine Learning (ICML)}, 2025.

\bibitem[Leng et~al.(2025)Leng, Singh, Hou, Xing, Xie, and
  Zheng]{leng2025repae}
Xingjian Leng, Jaskirat Singh, Yunzhong Hou, Zhenchang Xing, Saining Xie, and
  Liang Zheng.
\newblock {REPA-E}: Unlocking {VAE} for end-to-end tuning with latent diffusion
  transformers.
\newblock In \emph{IEEE/CVF International Conference on Computer Vision
  (ICCV)}, 2025.

\bibitem[Li et~al.(2025)Li, Tang, Xu, Wu, Zhou, Shao, Yu, Cao, and
  Lu]{li2025gamecraft}
Jiaqi Li, Junshu Tang, Zhiyong Xu, Longhuang Wu, Yuan Zhou, Shuai Shao, Tianbao
  Yu, Zhiguo Cao, and Qinglin Lu.
\newblock {Hunyuan-GameCraft}: High-dynamic interactive game video generation
  with hybrid history condition.
\newblock \emph{preprint arXiv:2506.17201}, 2025.

\bibitem[Liang et~al.(2025)Liang, Gojcic, Ling, Munkberg, Hasselgren, Lin, Gao,
  Keller, Vijaykumar, Fidler, and Wang]{liang2025diffusionrenderer}
Ruofan Liang, Zan Gojcic, Huan Ling, Jacob Munkberg, Jon Hasselgren, Chih-Hao
  Lin, Jun Gao, Alexander Keller, Nandita Vijaykumar, Sanja Fidler, and Zian
  Wang.
\newblock {DiffusionRenderer}: Neural inverse and forward rendering with video
  diffusion models.
\newblock In \emph{IEEE/CVF Conference on Computer Vision and Pattern
  Recognition (CVPR)}, 2025.

\bibitem[Lustig et~al.(2007)Lustig, Donoho, and Pauly]{lustig2007sparse}
Michael Lustig, David Donoho, and John~M. Pauly.
\newblock Sparse {MRI}: The application of compressed sensing for rapid {MR}
  imaging.
\newblock \emph{Magnetic Resonance in Medicine}, 58\penalty0 (6):\penalty0
  1182--1195, 2007.

\bibitem[Mardani et~al.(2024)Mardani, Song, Kautz, and
  Vahdat]{mardani2024reddiff}
Morteza Mardani, Jiaming Song, Jan Kautz, and Arash Vahdat.
\newblock A variational perspective on solving inverse problems with diffusion
  models.
\newblock In \emph{International Conference on Learning Representations
  (ICLR)}, 2024.

\bibitem[Martin~Garcia et~al.(2025)Martin~Garcia, Knaebel, Schmidt, de~Geus,
  Hermans, and Leibe]{garcia2025finetuning}
Gonzalo Martin~Garcia, Karim Knaebel, Christian Schmidt, Daan de Geus,
  Alexander Hermans, and Bastian Leibe.
\newblock Fine-tuning image-conditional diffusion models is easier than you
  think.
\newblock In \emph{IEEE/CVF Winter Conference on Applications of Computer
  Vision (WACV)}, 2025.

\bibitem[{NVIDIA}(2025)]{nvidia2025cosmos}
{NVIDIA}.
\newblock Cosmos world foundation model platform for physical {AI}.
\newblock \emph{preprint arXiv:2501.03575}, 2025.

\bibitem[Ong and Lustig(2019)]{ong2019sigpy}
Frank Ong and Michael Lustig.
\newblock {SigPy}: A {Python} package for high performance iterative
  reconstruction.
\newblock In \emph{ISMRM Annual Meeting \& Exhibition}, 2019.

\bibitem[Ozturkler et~al.(2023)Ozturkler, Liu, Eckart, Mardani, Song, and
  Kautz]{ozturkler2023smrd}
Batu Ozturkler, Chao Liu, Benjamin Eckart, Morteza Mardani, Jiaming Song, and
  Jan Kautz.
\newblock {SMRD}: {SURE}-based robust {MRI} reconstruction with diffusion
  models.
\newblock In \emph{International Conference on Medical Image Computing and
  Computer Assisted Intervention (MICCAI)}, 2023.

\bibitem[Pavlovic et~al.(2026)Pavlovic, Wandel, Obukhov, Bartolomei, Davydov,
  Tosi, Poggi, S{\"u}sstrunk, and Dai]{pavlovic2026marigoldv2}
Igor Pavlovic, Thiemo Wandel, Anton Obukhov, Luca Bartolomei, Andrey Davydov,
  Fabio Tosi, Matteo Poggi, Sabine S{\"u}sstrunk, and Dengxin Dai.
\newblock Marigold {V2}: Revisiting diffusion transformers for monocular depth
  estimation.
\newblock \emph{ACM Transactions on Graphics (SIGGRAPH Asia)}, 45\penalty0 (6),
  2026.

\bibitem[Perini et~al.(2026)Perini, Bischof, Arar, Duran, Kraus, Mishra, and
  Bickel]{perini2026windinet}
Janne Perini, Rafael Bischof, Moab Arar, Ay{\c{c}}a Duran, Michael~A. Kraus,
  Siddhartha Mishra, and Bernd Bickel.
\newblock Pretrained video models as differentiable physics simulators for
  urban wind flows.
\newblock In \emph{European Conference on Computer Vision (ECCV)}, 2026.

\bibitem[Poole et~al.(2023)Poole, Jain, Barron, and
  Mildenhall]{poole2023dreamfusion}
Ben Poole, Ajay Jain, Jonathan~T. Barron, and Ben Mildenhall.
\newblock {DreamFusion}: Text-to-{3D} using {2D} diffusion.
\newblock In \emph{International Conference on Learning Representations
  (ICLR)}, 2023.

\bibitem[Roberts et~al.(2021)Roberts, Ramapuram, Ranjan, Kumar, Bautista,
  Paczan, Webb, and Susskind]{roberts2021hypersim}
Mike Roberts, Jason Ramapuram, Anurag Ranjan, Atulit Kumar, Miguel~Angel
  Bautista, Nathan Paczan, Russ Webb, and Joshua~M. Susskind.
\newblock Hypersim: A photorealistic synthetic dataset for holistic indoor
  scene understanding.
\newblock In \emph{IEEE/CVF International Conference on Computer Vision
  (ICCV)}, 2021.

\bibitem[Romano et~al.(2017)Romano, Elad, and Milanfar]{romano2017red}
Yaniv Romano, Michael Elad, and Peyman Milanfar.
\newblock The little engine that could: Regularization by denoising ({RED}).
\newblock \emph{SIAM Journal on Imaging Sciences}, 10\penalty0 (4):\penalty0
  1804--1844, 2017.

\bibitem[Rombach et~al.(2022)Rombach, Blattmann, Lorenz, Esser, and
  Ommer]{rombach2022ldm}
Robin Rombach, Andreas Blattmann, Dominik Lorenz, Patrick Esser, and Bj{\"o}rn
  Ommer.
\newblock High-resolution image synthesis with latent diffusion models.
\newblock In \emph{IEEE/CVF Conference on Computer Vision and Pattern
  Recognition (CVPR)}, 2022.

\bibitem[Sch{\"o}ps et~al.(2017)Sch{\"o}ps, Sch{\"o}nberger, Galliani, Sattler,
  Schindler, Pollefeys, and Geiger]{schops2017eth3d}
Thomas Sch{\"o}ps, Johannes~L. Sch{\"o}nberger, Silvano Galliani, Torsten
  Sattler, Konrad Schindler, Marc Pollefeys, and Andreas Geiger.
\newblock A multi-view stereo benchmark with high-resolution images and
  multi-camera videos.
\newblock In \emph{IEEE Conference on Computer Vision and Pattern Recognition
  (CVPR)}, 2017.

\bibitem[Silberman et~al.(2012)Silberman, Hoiem, Kohli, and
  Fergus]{silberman2012nyu}
Nathan Silberman, Derek Hoiem, Pushmeet Kohli, and Rob Fergus.
\newblock Indoor segmentation and support inference from {RGBD} images.
\newblock In \emph{European Conference on Computer Vision (ECCV)}, 2012.

\bibitem[Sim{\'e}oni et~al.(2026)Sim{\'e}oni, Vo, Seitzer, Baldassarre, Oquab,
  et~al.]{simeoni2025dinov3}
Oriane Sim{\'e}oni, Huy~V. Vo, Maximilian Seitzer, Federico Baldassarre, Maxime
  Oquab, et~al.
\newblock {DINOv3}.
\newblock \emph{Transactions on Machine Learning Research (TMLR)}, 2026.

\bibitem[Skorokhodov et~al.(2025)Skorokhodov, Girish, Hu, Menapace, Li, Abdal,
  Tulyakov, and Siarohin]{skorokhodov2025diffusability}
Ivan Skorokhodov, Sharath Girish, Benran Hu, Willi Menapace, Yanyu Li, Rameen
  Abdal, Sergey Tulyakov, and Aliaksandr Siarohin.
\newblock Improving the diffusability of autoencoders.
\newblock In \emph{International Conference on Machine Learning (ICML)}, 2025.

\bibitem[Song et~al.(2024)Song, Kwon, Zhang, Hu, Qu, and
  Shen]{song2024resample}
Bowen Song, Soo~Min Kwon, Zecheng Zhang, Xinyu Hu, Qing Qu, and Liyue Shen.
\newblock Solving inverse problems with latent diffusion models via hard data
  consistency.
\newblock In \emph{International Conference on Learning Representations
  (ICLR)}, 2024.

\bibitem[Song and Ermon(2019)]{song2019ncsn}
Yang Song and Stefano Ermon.
\newblock Generative modeling by estimating gradients of the data distribution.
\newblock In \emph{Advances in Neural Information Processing Systems
  (NeurIPS)}, 2019.

\bibitem[Song et~al.(2021)Song, Sohl-Dickstein, Kingma, Kumar, Ermon, and
  Poole]{song2021sde}
Yang Song, Jascha Sohl-Dickstein, Diederik~P. Kingma, Abhishek Kumar, Stefano
  Ermon, and Ben Poole.
\newblock Score-based generative modeling through stochastic differential
  equations.
\newblock In \emph{International Conference on Learning Representations
  (ICLR)}, 2021.

\bibitem[Vahdat et~al.(2021)Vahdat, Kreis, and Kautz]{vahdat2021lsgm}
Arash Vahdat, Karsten Kreis, and Jan Kautz.
\newblock Score-based generative modeling in latent space.
\newblock In \emph{Advances in Neural Information Processing Systems
  (NeurIPS)}, 2021.

\bibitem[Valevski et~al.(2025)Valevski, Leviathan, Arar, and
  Fruchter]{valevski2025gamengen}
Dani Valevski, Yaniv Leviathan, Moab Arar, and Shlomi Fruchter.
\newblock Diffusion models are real-time game engines.
\newblock In \emph{International Conference on Learning Representations
  (ICLR)}, 2025.

\bibitem[Vasiljevic et~al.(2019)Vasiljevic, Kolkin, Zhang, Luo, Wang, Dai,
  Daniele, Mostajabi, Basart, Walter, and Shakhnarovich]{vasiljevic2019diode}
Igor Vasiljevic, Nick Kolkin, Shanyi Zhang, Ruotian Luo, Haochen Wang,
  Falcon~Z. Dai, Andrea~F. Daniele, Mohammadreza Mostajabi, Steven Basart,
  Matthew~R. Walter, and Gregory Shakhnarovich.
\newblock {DIODE}: A dense indoor and outdoor {DE}pth dataset.
\newblock \emph{preprint arXiv:1908.00463}, 2019.

\bibitem[Wan(2025)]{wan22}
Team Wan.
\newblock {Wan2.2}.
\newblock \url{https://github.com/Wan-Video/Wan2.2}, 2025.

\bibitem[Wan et~al.(2025)Wan, Wang, Ai, Wen, Mao, Xie, Chen, Yu, Zhao, Yang,
  et~al.]{wan2025wan}
Team Wan, Ang Wang, Baole Ai, Bin Wen, Chaojie Mao, Chen-Wei Xie, Di Chen,
  Feiwu Yu, Haiming Zhao, Jianxiao Yang, et~al.
\newblock Wan: Open and advanced large-scale video generative models.
\newblock \emph{preprint arXiv:2503.20314}, 2025.

\bibitem[Wang et~al.(2023{\natexlab{a}})Wang, Du, Li, Yeh, and
  Shakhnarovich]{wang2023sjc}
Haochen Wang, Xiaodan Du, Jiahao Li, Raymond~A. Yeh, and Greg Shakhnarovich.
\newblock {Score Jacobian Chaining}: Lifting pretrained {2D} diffusion models
  for {3D} generation.
\newblock In \emph{IEEE/CVF Conference on Computer Vision and Pattern
  Recognition (CVPR)}, 2023{\natexlab{a}}.

\bibitem[Wang et~al.(2026)Wang, Lin, Sun, Liu, Nie, Li, Liao, and
  Chu]{wang2026fe2e}
Jiyuan Wang, Chunyu Lin, Lei Sun, Rongying Liu, Lang Nie, Mingxing Li, Kang
  Liao, and Xiangxiang Chu.
\newblock {FE2E}: From editor to dense geometry estimator.
\newblock In \emph{IEEE/CVF Conference on Computer Vision and Pattern
  Recognition (CVPR)}, 2026.

\bibitem[Wang et~al.(2025)Wang, Xu, Dong, Deng, Xiang, Lv, Sun, Tong, and
  Yang]{wang2025moge2}
Ruicheng Wang, Sicheng Xu, Yue Dong, Yu Deng, Jianfeng Xiang, Zelong Lv,
  Guangzhong Sun, Xin Tong, and Jiaolong Yang.
\newblock {MoGe-2}: Accurate monocular geometry with metric scale and sharp
  details.
\newblock In \emph{Advances in Neural Information Processing Systems
  (NeurIPS)}, 2025.

\bibitem[Wang et~al.(2023{\natexlab{b}})Wang, Lu, Wang, Bao, Li, Su, and
  Zhu]{wang2023prolificdreamer}
Zhengyi Wang, Cheng Lu, Yikai Wang, Fan Bao, Chongxuan Li, Hang Su, and Jun
  Zhu.
\newblock {ProlificDreamer}: High-fidelity and diverse text-to-{3D} generation
  with variational score distillation.
\newblock In \emph{Advances in Neural Information Processing Systems
  (NeurIPS)}, 2023{\natexlab{b}}.

\bibitem[Wehenkel and Louppe(2021)]{wehenkel2021diffusionpriors}
Antoine Wehenkel and Gilles Louppe.
\newblock Diffusion priors in variational autoencoders.
\newblock In \emph{ICML Workshop on Invertible Neural Networks, Normalizing
  Flows, and Explicit Likelihood Models}, 2021.

\bibitem[Wu et~al.(2025)Wu, Li, Zhou, Lin, Gao, et~al.]{wu2025qwenimage}
Chenfei Wu, Jiahao Li, Jingren Zhou, Junyang Lin, Kaiyuan Gao, et~al.
\newblock {Qwen-Image} technical report.
\newblock \emph{preprint arXiv:2508.02324}, 2025.

\bibitem[Xu et~al.(2025)Xu, Ge, Liu, Fan, Xie, Zhao, Chen, and
  Shen]{xu2025genpercept}
Guangkai Xu, Yongtao Ge, Mingyu Liu, Chengxiang Fan, Kangyang Xie, Zhiyue Zhao,
  Hao Chen, and Chunhua Shen.
\newblock What matters when repurposing diffusion models for general dense
  perception tasks?
\newblock In \emph{International Conference on Learning Representations
  (ICLR)}, 2025.

\bibitem[Yao et~al.(2025)Yao, Yang, and Wang]{yao2025vavae}
Jingfeng Yao, Bin Yang, and Xinggang Wang.
\newblock Reconstruction vs.\ generation: Taming optimization dilemma in latent
  diffusion models.
\newblock In \emph{IEEE/CVF Conference on Computer Vision and Pattern
  Recognition (CVPR)}, 2025.

\bibitem[Ye et~al.(2024)Ye, Qiu, Gu, Zuo, Wu, Dong, Bo, Xiu, and
  Han]{ye2024stablenormal}
Chongjie Ye, Lingteng Qiu, Xiaodong Gu, Qi Zuo, Yushuang Wu, Zilong Dong,
  Liefeng Bo, Yuliang Xiu, and Xiaoguang Han.
\newblock {StableNormal}: Reducing diffusion variance for stable and sharp
  normal.
\newblock \emph{ACM Transactions on Graphics (SIGGRAPH Asia)}, 43\penalty0 (6),
  2024.

\bibitem[Yin et~al.(2024)Yin, Gharbi, Zhang, Shechtman, Durand, Freeman, and
  Park]{yin2024dmd}
Tianwei Yin, Micha{\"e}l Gharbi, Richard Zhang, Eli Shechtman, Fr{\'e}do
  Durand, William~T. Freeman, and Taesung Park.
\newblock One-step diffusion with distribution matching distillation.
\newblock In \emph{IEEE/CVF Conference on Computer Vision and Pattern
  Recognition (CVPR)}, 2024.

\bibitem[Yu et~al.(2025)Yu, Kwak, Jang, Jeong, Huang, Shin, and
  Xie]{yu2025repa}
Sihyun Yu, Sangkyung Kwak, Huiwon Jang, Jongheon Jeong, Jonathan Huang, Jinwoo
  Shin, and Saining Xie.
\newblock Representation alignment for generation: Training diffusion
  transformers is easier than you think.
\newblock In \emph{International Conference on Learning Representations
  (ICLR)}, 2025.

\bibitem[Zbontar et~al.(2018)Zbontar, Knoll, Sriram, Murrell, Huang, Muckley,
  Defazio, Stern, Johnson, Bruno, et~al.]{zbontar2018fastmri}
Jure Zbontar, Florian Knoll, Anuroop Sriram, Tullie Murrell, Zhengnan Huang,
  Matthew~J. Muckley, Aaron Defazio, Ruben Stern, Patricia Johnson, Mary Bruno,
  et~al.
\newblock {fastMRI}: An open dataset and benchmarks for accelerated {MRI}.
\newblock \emph{preprint arXiv:1811.08839}, 2018.

\bibitem[Zeng et~al.(2024)Zeng, Deschaintre, Georgiev, Hold-Geoffroy, Hu, Luan,
  Yan, and Ha{\v{s}}an]{zeng2024rgbx}
Zheng Zeng, Valentin Deschaintre, Iliyan Georgiev, Yannick Hold-Geoffroy, Yiwei
  Hu, Fujun Luan, Ling-Qi Yan, and Milo{\v{s}} Ha{\v{s}}an.
\newblock {RGB}$\leftrightarrow${X}: Image decomposition and synthesis using
  material- and lighting-aware diffusion models.
\newblock In \emph{ACM SIGGRAPH Conference Papers}, 2024.

\end{thebibliography}
}

\clearpage
\ifdefined\arxiv\else\setcounter{page}{1}\fi   %
\maketitlesupplementary

\setcounter{section}{0}
\renewcommand{\thesection}{\Alph{section}}
\renewcommand{\theHsection}{suppl.\Alph{section}}
\renewcommand{\dblfloatpagefraction}{0.5}   %
\makeatletter
\newcommand{\needlines}[1]{\par\penalty-100\begingroup\setlength{\dimen@}{#1\baselineskip}\dimen@ii\pagegoal
  \advance\dimen@ii-\pagetotal\ifdim\dimen@>\dimen@ii\ifdim\dimen@ii>\z@\vfil\fi\break\fi\endgroup}
\makeatother

This supplement follows the structure of the main paper.
\Cref{sec:suppl_gain} defines the gains and round-trip errors of \cref{fig:teaser}, \cref{sec:suppl_method} details the anchor of \cref{sec:method}, and \cref{sec:suppl_normals,sec:suppl_ablation,sec:suppl_latent,sec:suppl_modify,sec:suppl_prune,sec:suppl_sota,sec:suppl_mri,sec:suppl_wind,sec:suppl_cost} give the training details and additional results of \cref{sec:exp_normals,sec:exp_ablation,sec:exp_latent,sec:exp_widen,sec:exp_prune,sec:exp_sota,sec:exp_mri,sec:exp_wind,sec:exp_cost}.
Depth and normals are evaluated with the code and protocols of Marigold~V2~\cite{pavlovic2026marigoldv2}, including its RANSAC alignment in log space for depth, so that \cref{tab:final} compares with its published numbers.
For surface normals, the indoor error is the mean angular error over iBims-1, NYUv2, and ScanNet, the three benchmarks of \cref{tab:normals}.

\section{Gains across tasks}
\label{sec:suppl_gain}
\Cref{fig:teaser} plots the relative gain of DPA over the control in each task against the round-trip error of the frozen VAE in that task.
The gain is the relative reduction of the error metric of the task, namely AbsRel for depth with one VAE per modality (\cref{tab:widen}, 3.3\%), the mean angular error for normals (\cref{tab:normals}, 4.0\%), the test VRMSE for wind flows with LTX-Video (\cref{tab:wind}, 5.2\%), and for MRI the reduction of the RMSE of the magnitude image implied by the PSNR gain of 0.65\,dB (\cref{tab:mri}, 7.2\%).
The round-trip error is a normalized RMSE, so that it compares across modalities.
Per sample and target channel, the mean squared error over the valid pixels is divided by the variance of the target, the result is averaged over channels, and its square root is averaged over samples.
Channels without variance, such as the zero channel of MRI, are skipped.
We measure the round-trip error on the encoder means, decoded without the noise floor, in the target space of each task, on 200 Hypersim validation frames for depth and normals, 100 validation slices for MRI, and 100 of the 1{,}000 test simulations of WinDiNet for wind flows.
The four-channel VAE has no frozen counterpart, so \cref{fig:teaser} shows its depth gain (9.3\%, \cref{tab:widen}) without a round-trip error.

\section{Method details}
\label{sec:suppl_method}

\paragraph{Pretraining distribution.}
The pretraining distribution $\pi$ is the shifted logit-normal with which the diffusion model is fine-tuned, $\lambda\sim\mathcal{N}(-2\log s,2^2)$, truncated to $[\lambda_\mathrm{min},\lambda_\mathrm{max}]$, with the shift $s=6.93$ for FLUX~3.

\paragraph{Anchor gradient.}
With $w=1/\sigma_\lambda$ (\cref{sec:adjust}), a draw from $\pi$ gives the velocity residual $\hat v_\phi-v=(\hat\epsilon_\phi-\epsilon)/\alpha_\lambda$ with target $v=\epsilon-z$, which is the residual of the velocity loss $\tfrac{1}{2}\|\hat v_\phi-v\|^2$ on which the diffusion model is trained, and its gradient through the target alone.
The uniform component covers the low noise levels that $\pi$ rarely draws.
With the anchor averaged over the latent elements, $\beta=1$ gives an anchor gradient of 10--16\% of the norm of the reconstruction gradient on the pretrained FLUX~3 VAE (median 15\%, 8 batches of 16 Hypersim frames).
Most of the anchor gradient comes from the levels drawn from $\pi$, and it is nearly orthogonal to the reconstruction gradient (cosine $-0.03$).

\paragraph{Noise floor.}
The log-SNR range $[\lambda_\mathrm{min},\lambda_\mathrm{max}]=[-8,4]$ sets the noise floor to $\sigma_0=e^{-2}\approx0.135$ in normalized latent units, the noise level at the top of the anchor's range.
Only VAEs trained with the anchor use the noise floor.

\paragraph{DiT loss.}
The DiT loss of \cref{sec:setup} is the velocity loss $\tfrac{1}{2}\|\hat v_\phi-v\|^2$ of the frozen diffusion model at its training noise levels $\pi$.
We evaluate it per latent element on the encoder means of 16 validation images at the 8 quantiles of $\pi$ with probabilities $(k+\tfrac{1}{2})/8$, with fixed noise and the unconditional prediction.

\section{Surface normal estimation}
\label{sec:suppl_normals}

\paragraph{Training.}
The VAE is fine-tuned for one epoch (about 2{,}500 steps) at batch size 16 and learning rate $10^{-4}$, with 50 warm-up steps, gradient clipping at 10, the L1 reconstruction loss, and the KL weight $10^{-6}$.
FLUX~3 Image Klein Base is then fine-tuned for two epochs (about 5{,}000 steps) at batch size 16 and learning rate $10^{-4}$, decayed linearly to $10^{-5}$, with a fixed prompt (``Turn this photo into a surface normal map.'') and the text path frozen.
Both stages use the normal maps at $576\times768$.

\begin{figure*}[p]
  \centering
  \includegraphics[width=\linewidth]{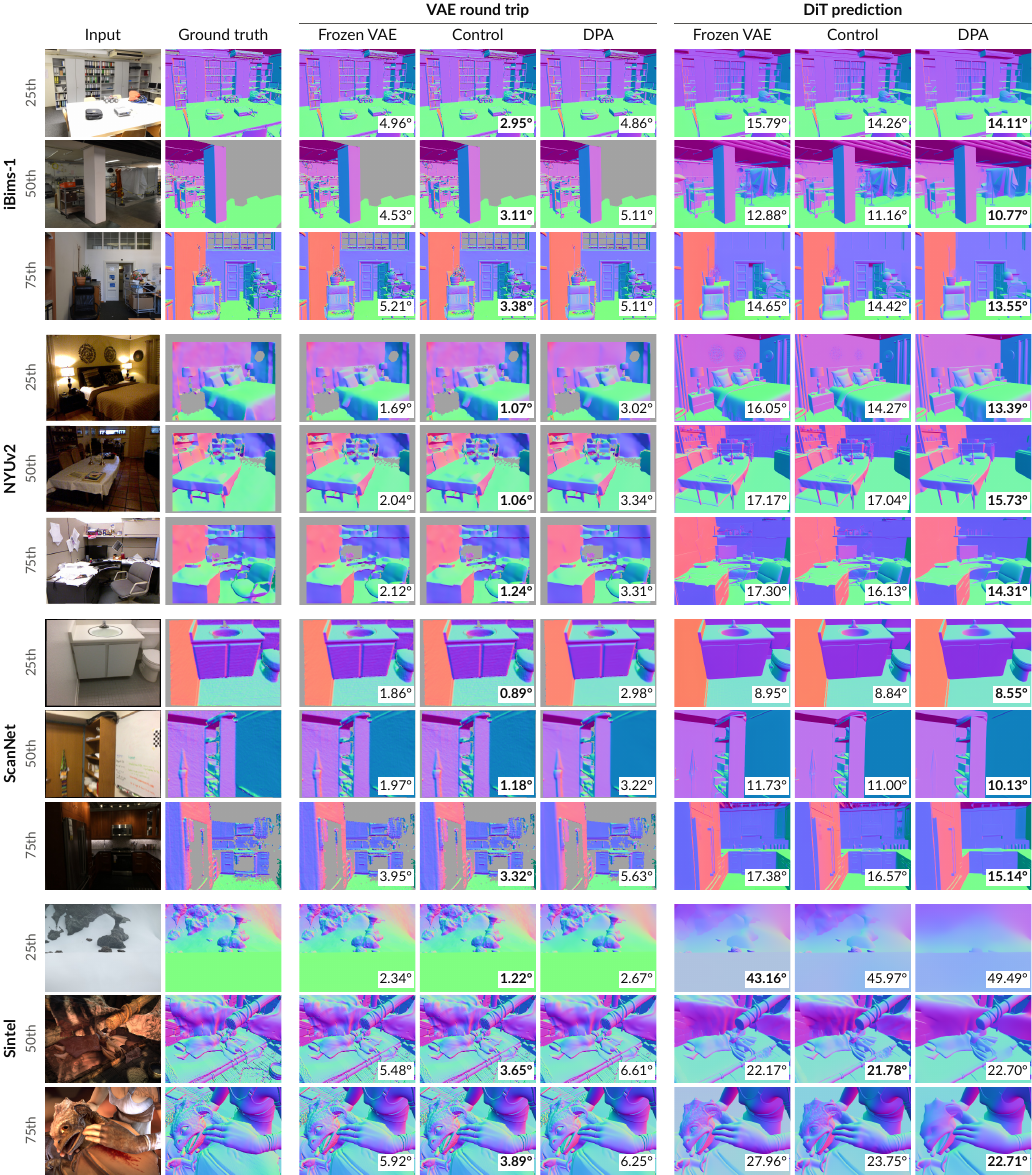}
  \caption{\textbf{Qualitative surface normals at fixed quantiles.} On each benchmark, the test images at the 25th, 50th, and 75th percentile of the per-image gain of DPA over the control. Round trip: the ground truth encoded and decoded by each VAE. Prediction: FLUX~3 fine-tuned on the latents of each VAE, sampled with 4 steps. Numbers: mean angular error over the image (degrees), the lowest per group in bold.}
  \label{fig:suppl_normals_quartiles}
\end{figure*}

\paragraph{Qualitative results.}
\Cref{fig:normals_qual} shows a selected image, and \cref{fig:suppl_normals_quartiles} shows images chosen by a fixed rule, namely the test images at the 25th, 50th, and 75th percentile of the per-image gain of DPA over the control on each benchmark.
The model trained on DPA latents gives the most accurate prediction on all nine indoor images shown.
As in \cref{tab:normals}, this holds despite a higher round-trip error than that of the control.

\paragraph{Number of sampling steps.}
\Cref{tab:suppl_steps} evaluates the models of \cref{tab:normals} with 1 to 16 Euler steps instead of two.
Two steps give the lowest error for both the control and DPA.
With more steps, the error of both models increases, faster for the control, so the advantage of DPA on the indoor benchmarks grows from $0.63^\circ$ at two steps to $1.02^\circ$ at 16.

\begin{table}[t]
  \centering
  \small
  \caption{\textbf{Number of sampling steps on normals.} Mean angular error (degrees) of the models of \cref{tab:normals} with $n$ Euler steps.}
  \label{tab:suppl_steps}
  \resizebox{\ifdim\width>\columnwidth\columnwidth\else\width\fi}{!}{%
  \begin{tabular}{@{}lcccccc@{}}
    \toprule
    & \multicolumn{3}{c}{Indoor$\downarrow$} & \multicolumn{3}{c}{Sintel$\downarrow$} \\
    \cmidrule(lr){2-4}\cmidrule(l){5-7}
    Steps & Control & DPA & DPA $-$ control & Control & DPA & DPA $-$ control \\
    \midrule
    1 & 16.08 & 15.32 & $-0.76$ & 28.30 & 30.51 & $+2.21$ \\
    2 & \textbf{15.61} & \textbf{14.98} & $-0.63$ & \textbf{28.04} & \textbf{30.03} & $+1.99$ \\
    4 & 16.10 & 15.19 & $-0.91$ & 28.64 & 30.82 & $+2.18$ \\
    8 & 16.57 & 15.57 & $-1.00$ & 29.67 & 31.76 & $+2.08$ \\
    16 & 16.94 & 15.93 & $-1.02$ & 30.43 & 32.46 & $+2.03$ \\
    \bottomrule
  \end{tabular}%
  }
\end{table}

\needlines{8}
\paragraph{Test-time ensembling.}
Since the diffusion model samples its prediction from noise, several samples can be averaged.
We take the mean of their unit normals and renormalize it.
\Cref{tab:suppl_avg} gives the indoor error with one to four samples, for fewer than four averaged over all subsets of the four.
The control benefits more from averaging than DPA, so the advantage of DPA narrows with more samples, from $0.77^\circ$ with one sample to $0.52^\circ$ with four on Hypersim, and from $0.68^\circ$ to $0.48^\circ$ for the models trained with Virtual KITTI~2 (\cref{sec:suppl_sota}).
DPA with one sample remains more accurate than the control with four ($14.97^\circ$ vs.\ $15.37^\circ$ on Hypersim).
On Sintel, DPA trained on Hypersim is about $2.2^\circ$ less accurate than the control for every number of samples.

\begin{table}[h]
  \centering
  \small
  \caption{\textbf{Test-time ensembling.} Indoor mean angular error (degrees) with one to four samples averaged. For fewer than four, the mean over all subsets of four samples. $\Delta$: DPA $-$ control.}
  \label{tab:suppl_avg}
  \resizebox{\columnwidth}{!}{%
  \begin{tabular}{@{}ccccccc@{}}
    \toprule
    & \multicolumn{3}{c}{Hypersim} & \multicolumn{3}{c}{Hypersim and Virtual KITTI~2} \\
    \cmidrule(lr){2-4}\cmidrule(l){5-7}
    Samples & Control & DPA & $\Delta$ & Control & DPA & $\Delta$ \\
    \midrule
    1 & 15.73 & 14.97 & $-0.77$ & 15.55 & 14.87 & $-0.68$ \\
    2 & 15.50 & 14.89 & $-0.60$ & 15.39 & 14.85 & $-0.55$ \\
    3 & 15.41 & 14.87 & $-0.54$ & 15.34 & 14.84 & $-0.50$ \\
    4 & 15.37 & 14.85 & $-0.52$ & 15.31 & 14.83 & $-0.48$ \\
    \bottomrule
  \end{tabular}}
\end{table}

\paragraph{VAE batch size.}
The VAEs are fine-tuned at batch size 16.
To test whether a larger batch changes the comparison, we fine-tune the control and the DPA VAE at batch size 64 for the same number of steps, that is, for four epochs at four times the compute, and then the diffusion model on each as in \cref{tab:normals}.
On the indoor benchmarks, the control then reaches $15.69^\circ$ ($15.61^\circ$ at batch size 16) and DPA $14.96^\circ$ ($14.98^\circ$), so DPA is $0.73^\circ$ more accurate than the control, against $0.63^\circ$ at batch size 16.
The DPA latents differ from those at batch size 16 by 3.8\% (relative $L_2$ distance on the validation frames).
A larger batch thus does not improve either model enough to justify four times the compute, so we keep batch size 16.
Note that the gain of DPA over the control grows with the larger batch.

\begin{table}[t]
  \centering
  \small
  \caption{\textbf{Surface normals with Qwen-Image-Edit-2509.} Mean angular error (degrees). Rec.\ and Mean as in \cref{tab:normals}. Decoder-only keeps the latents of the frozen encoder and uses the diffusion model of the frozen VAE.}
  \label{tab:suppl_qwen}
  \resizebox{\columnwidth}{!}{%
  \begin{tabular}{@{}lcccccc@{}}
    \toprule
    VAE adaptation & Rec.$\downarrow$ & iBims-1$\downarrow$ & NYUv2$\downarrow$ & ScanNet$\downarrow$ & Mean$\downarrow$ & Sintel$\downarrow$ \\
    \midrule
    Frozen & 2.98 & 15.79 & \textbf{16.10} & \textbf{14.54} & \textbf{15.48} & 33.94 \\
    Decoder-only & 2.15 & 15.81 & 16.26 & 14.61 & 15.56 & 33.83 \\
    Control & \textbf{2.04} & 15.92 & 16.39 & 14.93 & 15.75 & \textbf{31.72} \\
    DPA ($\beta=0.5$) & 2.34 & \textbf{15.56} & 16.52 & 14.79 & 15.62 & 31.87 \\
    \bottomrule
  \end{tabular}%
  }
\end{table}

\begin{table}[t]
  \centering
  \small
  \caption{\textbf{Further variants of DPA} (the others are in \cref{tab:weight}). Columns as in \cref{tab:weight}. $\Delta$: difference to DPA, $^\dagger$ to the row above.}
  \label{tab:ablation}
  \resizebox{\columnwidth}{!}{%
  \begin{tabular}{@{}lcccc@{}}
    \toprule
    Variant & Rec.$\downarrow$ & Std & Normals$\downarrow$ & $\Delta$ \\
    \midrule
    DPA ($\beta=1$, mixed levels, $\sigma_0=0.135$, no condition) & 2.89 & 0.55 & \textbf{14.98} & -- \\
    \midrule
    Mixed levels, residual not divided by $\alpha_\lambda$ & 2.14 & 0.73 & 15.27 & $+0.29$ \\
    Prior conditioned on the task prompt & 2.94 & 0.54 & 15.09 & $+0.11$ \\
    Noise floor $\sigma_0=0.22$ & 3.65 & 0.55 & 15.03 & $+0.05$ \\
    \midrule
    Angular reconstruction loss & 2.33 & 0.55 & 15.03 & $+0.05$ \\
    \quad with VSD~\cite{wang2023prolificdreamer} & 1.70 & 0.78 & 15.74 & $+0.71^\dagger$ \\
    \midrule
    Trajectory adjustment of FLAIR~\cite{erbach2025flair} & 2.77 & 0.55 & 15.07 & $+0.09$ \\
    \bottomrule
  \end{tabular}%
  }
\end{table}

\paragraph{Second image backbone.}
We repeat the normals experiment with Qwen-Image-Edit-2509~\cite{wu2025qwenimage}, on which Marigold~V2 is built, with the conditioning, prompt, and sampler of FLUX~3.
Its diffusion model is fine-tuned with LoRA of rank 128 on its image stream for as long as FLUX~3 trains (about 1{,}850 steps, \cref{sec:suppl_cost}), and DPA uses $\beta=0.5$ with its noise levels shifted to the noise schedule of Qwen-Image.
On this backbone too, DPA outperforms the control, by $0.12^\circ$ on the indoor benchmarks (\cref{tab:suppl_qwen}).

\section{Ablation studies}
\label{sec:suppl_ablation}

\paragraph{Design choices.}
\Cref{tab:ablation} lists variants of DPA beyond those of \cref{tab:weight}.
Without the division by $\alpha_\lambda$ for draws from $\pi$, the mixture of noise levels increases the error by $0.29^\circ$.
With the range of the anchor unchanged, a noise floor of $\sigma_0=0.22$, 1.6 times above the default, changes the error by $0.05^\circ$, as $\sigma_0=0.05$ does in \cref{tab:weight}.
A conditioned prior increases the error, by $0.11^\circ$ with the task prompt and by $0.52^\circ$ when the input image enters as reference, so we anchor to the unconditional prior.
Two further variants address the mode-seeking behavior of SDS.
Variational score distillation (VSD)~\cite{wang2023prolificdreamer}, which replaces the noise in the anchor with the prediction of a LoRA of rank 16 fitted online to the current latents, increases the error by $0.71^\circ$ relative to DPA with the same angular reconstruction loss.
The trajectory adjustment of FLAIR~\cite{erbach2025flair}, which replaces part of the noise by the prediction of the prior at a slightly higher noise level, increases the error by $0.09^\circ$.

\begin{figure*}[t]
  \centering
  \includegraphics[width=\linewidth]{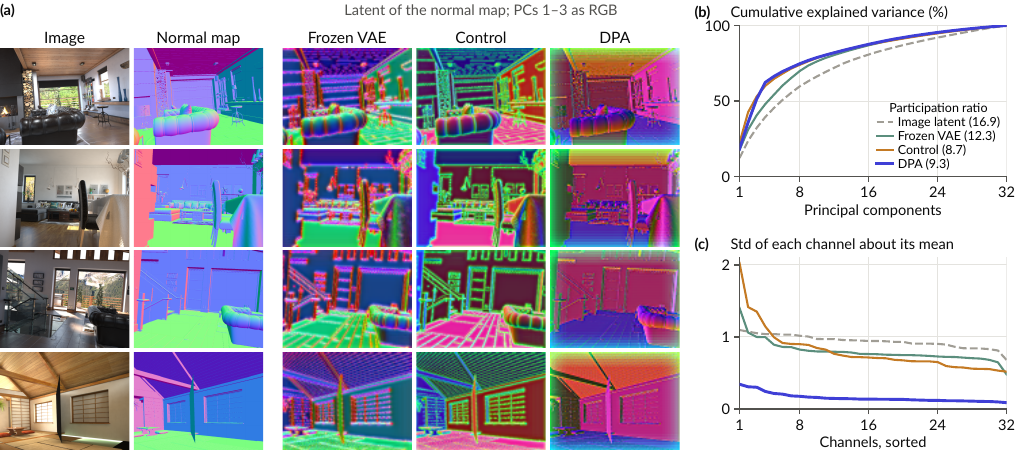}
  \caption{\textbf{Principal components of the latent channels.} Normalized latents (32 channels) on 64 Hypersim validation frames, with one PCA basis per VAE. (a)~Four frames: the RGB image, the ground-truth normal map, and the projection of the normal-map latents of the frozen, control, and DPA VAEs on their first three principal components as RGB. (b)~Cumulative explained variance, with the participation ratio $(\sum_k\lambda_k)^2/\sum_k\lambda_k^2$, an effective dimensionality, in parentheses. (c)~Standard deviation of each channel about its mean, sorted.}
  \label{fig:suppl_latent_pca}
\end{figure*}

\begin{figure*}[t]
  \centering
  \includegraphics[width=\linewidth]{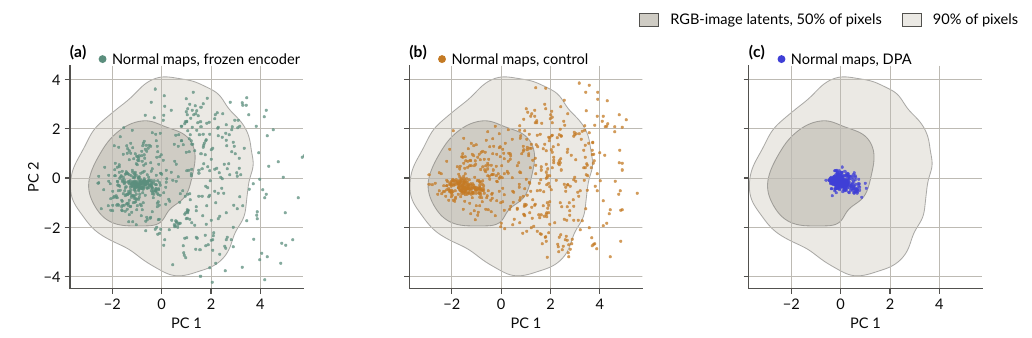}
  \caption{\textbf{Distribution of the latent pixels relative to the image latents.} Latent pixels (32-dimensional) of 64 Hypersim validation frames, projected on the first two principal components of the frozen VAE's latents of RGB images, centered on their mean. Shaded: the regions holding 50\% and 90\% of the RGB-image latent pixels. Dots: 600 random latent pixels of the normal maps under (a)~the frozen encoder, (b)~the control, and (c)~DPA.}
  \label{fig:suppl_latent_2d}
\end{figure*}

\paragraph{Data efficiency.}
The subsets of \cref{fig:data} hold 394, 3{,}942, and 9{,}854 frames.
Both stages train for 10\%, 32\%, and 50\% of the steps used with all frames, the square root of the share of the frames.
From all frames to 1\%, the round-trip errors rise from $1.54^\circ$ to $2.01^\circ$ for the control and from $2.89^\circ$ to $3.17^\circ$ for DPA.

\section{Latent space analysis}
\label{sec:suppl_latent}

\begin{table}[t]
  \centering
  \small
  \caption{\textbf{Alternatives to the anchor on normals.} Latent anchoring: L2 penalty toward the latents of the frozen encoder, with weight 1 or 0.1. FLOPs-matched: the control with 4{,}600 VAE steps, the FLOPs of DPA. Rec., Std, Dist., and DiT loss as in \cref{tab:weight}, errors in degrees. Indoor: iBims-1, NYUv2, and ScanNet. All: with Sintel. $\Delta$: indoor difference to DPA.}
  \label{tab:suppl_alternatives}
  \resizebox{\ifdim\width>\columnwidth\columnwidth\else\width\fi}{!}{%
  \begin{tabular}{@{}lccccrrrr@{}}
    \toprule
    VAE & Rec.$\downarrow$ & Std & Dist. & DiT loss$\downarrow$ & Indoor$\downarrow$ & Sintel$\downarrow$ & All$\downarrow$ & $\Delta$ \\
    \midrule
    Frozen & 3.54 & 0.83 & 0.00 & 0.136 & 16.19 & 29.13 & 19.43 & $+1.21$ \\
    Decoder-only & 2.16 & 0.83 & 0.00 & 0.136 & 15.75 & 28.47 & 18.93 & $+0.77$ \\
    Latent anchoring, 1 & 1.72 & 0.83 & 0.02 & 0.136 & 15.73 & 28.48 & 18.91 & $+0.74$ \\
    Latent anchoring, 0.1 & 1.63 & 0.83 & 0.07 & 0.134 & 15.70 & 27.91 & 18.76 & $+0.72$ \\
    Control & 1.54 & 0.85 & 0.69 & 0.190 & 15.61 & 28.04 & 18.72 & $+0.63$ \\
    Control, rescaled & 1.54 & 0.55 & 0.64 & 0.117 & 15.69 & 28.14 & 18.80 & $+0.70$ \\
    Control, FLOPs-matched & \textbf{1.42} & 0.86 & 0.77 & 0.199 & 15.71 & 29.12 & 19.06 & $+0.72$ \\
    Control, angular loss & 1.49 & 0.86 & 0.71 & 0.196 & 15.66 & \textbf{26.65} & \textbf{18.40} & $+0.67$ \\
    VA-VAE alignment & 1.83 & 1.86 & 2.28 & 0.977 & 16.18 & 28.32 & 19.21 & $+1.19$ \\
    \midrule
    DPA & 2.89 & 0.55 & 1.00 & \textbf{0.036} & \textbf{14.98} & 30.03 & 18.75 & -- \\
    \bottomrule
  \end{tabular}%
  }
\end{table}

\paragraph{Principal component analysis.}
DPA reduces the standard deviation of the latents from 0.85 for the control to 0.55 (\cref{tab:weight}).
A smaller standard deviation can arise in two ways: the latents can occupy fewer directions of the latent space, or they can vary less within each direction.
\Cref{fig:suppl_latent_pca} distinguishes the two.
If DPA concentrated the latents in fewer directions, fewer principal components would hold most of their variance, and its curve in \cref{fig:suppl_latent_pca}b would rise faster than that of the control.
Instead, the two curves nearly coincide: the DPA and control latents both need 19 of the 32 components for 90\% of their variance (the image latents 23), and they have about the same effective dimensionality (participation ratio 9.3 and 8.7).
DPA therefore keeps the dimensionality of the latents and reduces their variation along all directions.
Each channel varies about a fifth as much around its own mean as in the control (0.17 vs.\ 0.86, \cref{fig:suppl_latent_pca}c), so the overall standard deviation of 0.55 consists mostly of offsets of the channel means (0.53).
\Cref{fig:suppl_latent_2d} shows where the latent pixels land relative to the frozen VAE's latents of RGB images, the distribution the frozen DiT is trained on, in the plane of its first two principal components.
The frozen encoder spreads the normal-map latents over this distribution and beyond it, and the control keeps this spread, whereas DPA maps all latent pixels into a region near its center that holds 3\% of the variance of the RGB-image latents (0.90 vs.\ 28.5).
This plane holds only 10\% of the variance of the DPA latents, but the same holds in all 32 channels: under a Gaussian fit to the RGB-image latents, the median DPA latent pixel lies closer to its mean than 88\% of the RGB-image latent pixels.
DPA thus encodes the normal map in the same number of directions at a smaller amplitude around shifted channel means, in a high-density region of the RGB-image distribution, consistent with the nearly uniform latents of flat surfaces in \cref{fig:normals_qual}.

\paragraph{Alternatives to the anchor.}
These properties of the DPA latents raise the question whether the gain comes from the anchor itself or from simpler properties that other changes to the VAE fine-tuning could also produce, such as a smaller scale of the latents.
\Cref{tab:suppl_alternatives} tests this with every variant we trained on normals without the anchor.
For each, it reports the reconstruction of the VAE (Rec.), statistics of its latents (Std, Dist., and the loss of the frozen DiT on them), and the downstream error after fine-tuning the DiT on these latents.

None of the variants lowers the downstream error below that of the control ($15.61^\circ$).
The most direct test of the scale is the rescaled control, which matches the standard deviation of the DPA latents by a global factor.
Its latents still vary 3.3 times as much around their channel means as the DPA latents (0.56 against 0.17), so they do not have the structure of the DPA latents, and its downstream error stays at that of the control ($15.69^\circ$).
Another variant regularizes the latents as VA-VAE~\cite{yao2025vavae} does, with the features of a vision foundation model (DINOv3~\cite{simeoni2025dinov3}) instead of a diffusion prior.
Its latents are 3.3 times as far from those of the frozen encoder as the control latents (Dist.\ 2.28 against 0.69), the frozen DiT has a five times higher loss on them (0.977 against 0.190), and its downstream indoor error is $0.56^\circ$ higher than that of the control.

\begin{table}[t]
  \centering
  \small
  \caption{\textbf{Rate by noise level.} Noise-prediction error of the frozen FLUX~3 per latent element at eight log-SNRs, and the rate $R$, on 64 Hypersim validation frames with shared noise. Rows: the frozen VAE on images, the VAEs of \cref{tab:normals} and DPA with uniform levels on normal maps, and Gaussian noise.}
  \label{tab:suppl_rate}
  \resizebox{\columnwidth}{!}{%
  \begin{tabular}{@{}lccccccccc@{}}
    \toprule
    $\lambda$ & $-7.25$ & $-5.75$ & $-4.25$ & $-2.75$ & $-1.25$ & $0.25$ & $1.75$ & $3.25$ & $R$ \\
    \midrule
    Frozen VAE, images & 0.000 & 0.001 & 0.003 & 0.010 & 0.030 & 0.086 & 0.197 & 0.336 & 1.00 \\
    \midrule
    Frozen & 0.000 & 0.001 & 0.002 & 0.005 & 0.013 & 0.035 & 0.083 & 0.161 & 0.45 \\
    Control & 0.000 & 0.001 & 0.003 & 0.007 & 0.021 & 0.057 & 0.133 & 0.248 & 0.71 \\
    DPA & 0.000 & 0.000 & 0.000 & 0.001 & 0.008 & 0.045 & 0.141 & 0.264 & 0.69 \\
    DPA, uniform levels & 0.000 & 0.000 & 0.001 & 0.002 & 0.005 & 0.012 & 0.030 & 0.072 & 0.19 \\
    \midrule
    Gaussian noise & 0.000 & 0.002 & 0.007 & 0.032 & 0.123 & 0.363 & 0.636 & 0.684 & 2.77 \\
    \bottomrule
  \end{tabular}%
  }
\end{table}

\paragraph{Likelihood under the prior.}
The DPA latents thus lie in a high-density region of the distribution of RGB-image latents (\cref{fig:suppl_latent_2d}).
To test whether they are also more likely under the prior itself, we measure the rate $R(z)=\tfrac{1}{2}\int_{\lambda_\mathrm{min}}^{\lambda_\mathrm{max}}\mathbb{E}_\epsilon\|\hat\epsilon_\phi(z_\lambda,\lambda)-\epsilon\|^2\,d\lambda$.
Up to two terms at $\lambda_\mathrm{min}$ and $\lambda_\mathrm{max}$, it is an upper bound on $-\log p_\phi(z)$~\cite{kingma2021vdm,kingma2023understanding}.

On the 64 validation frames of \cref{fig:suppl_latent_pca}, the frozen VAE's latents of the normal maps have the lowest rate of the three VAEs of \cref{tab:normals} (0.45), and the control and DPA have the same rate (0.71 and 0.69, DPA lower on 32 of the 64 frames).
The rate therefore does not follow the downstream errors (\cref{tab:suppl_rate}).
All three lie below the rate of the frozen VAE's latents of the RGB frames (1.00) and below that of Gaussian noise (2.77).
The two least noisy levels hold more than 80\% of the rate, and there the error on the DPA latents is 6\% higher than on the control latents (0.141 and 0.264 vs.\ 0.133 and 0.248, \cref{tab:suppl_rate}).
At every level up to $\lambda=0.25$, they are denoised better, by a factor of about 11 at $\lambda=-4.25$.
DPA with uniform noise levels, the weighting of this bound, lowers the rate to 0.19, the lowest of all rows, but is $0.35^\circ$ less accurate than DPA (\cref{tab:weight}).

\section{VAE channel expansion for multi-modal data}
\label{sec:suppl_modify}

\paragraph{Initialization.}
For targets with three channels, such as normals, the VAE keeps its pretrained input and output layers.
For $C\neq3$ channels in $K$ modalities (\eg, one channel for depth and three for normals), the first convolution of the encoder and the last convolution of the decoder are replaced by convolutions with $C$ channels, one group per modality.
Each group is initialized from the pretrained RGB layer: the encoder applies the pretrained kernel to every group and averages the $K$ responses, and every group of the decoder copies the pretrained RGB output.
A one-channel modality is encoded as an RGB image with three equal channels and decoded as the mean of the three RGB outputs.

\paragraph{Training.}
The reconstruction loss is the mean of the losses of the modalities, so that the weights of the KL term and the anchor keep their meaning for every $C$.
The anchor is off for the first 5\% of the steps and reaches $\beta=1$ linearly over the next 10\%, so that the new layers learn the target before the anchor regularizes the latents.
The four-channel VAEs weight the losses of depth and normals with GradNorm~\cite{chen2018gradnorm} (asymmetry 1.5), which balances the gradients of the modalities on the input of the last layer of the decoder, with weights of mean one.
The four-channel VAEs and the VAEs for one modality are fine-tuned for 10{,}000 steps on the 21{,}239 Hypersim frames shared by the depth and normal splits of Marigold~\cite{ke2025marigold} at $576\times768$, with 2{,}398 frames of other scenes for validation.
Log depth is normalized by its per-image percentiles, and normals are given in camera space~\cite{ke2025marigold}.
The diffusion model follows the normals recipe of \cref{sec:suppl_normals} with the multi-channel latent as its target.
Depth errors are means over four sampling seeds.

\paragraph{Tiling.}
An alternative to channel expansion keeps the frozen RGB VAE and encodes depth (replicated to three channels) and normals into two latents, which a single DiT predicts side by side as one latent of double width (Marigold~\cite{ke2025marigold} concatenates such latents along channels instead).
Trained and evaluated as in \cref{tab:widen}, tiling matches the four-channel DPA model on depth (AbsRel $\times100$ 6.01 vs.\ 6.02), but it is $1.73^\circ$ less accurate on normals ($16.78^\circ$ vs.\ $15.04^\circ$) and needs twice as many latent tokens.

\begin{table}[t]
  \centering
  \small
  \caption{\textbf{Pruned VAEs.} Removed: share of the VAE's parameters. Time: encoding and decoding time relative to the unpruned VAE. Rec.: round-trip error of the VAE. Error: as in \cref{tab:normals,tab:wind}.}
  \label{tab:suppl_prune}
  \resizebox{\ifdim\width>\columnwidth\columnwidth\else\width\fi}{!}{%
  \begin{tabular}{@{}cccccccc@{}}
    \toprule
    & & \multicolumn{2}{c}{Rec.$\downarrow$} & \multicolumn{2}{c}{Error$\downarrow$} & \\
    \cmidrule(lr){3-4}\cmidrule(lr){5-6}
    Removed & Time & Control & DPA & Control & DPA & DPA $-$ control \\
    \midrule
    \multicolumn{7}{@{}l}{\emph{Normals, FLUX~3: mean angular error (degrees)}} \\
    0\% & 1 & 1.54 & 2.89 & 15.61 & \textbf{14.98} & $-0.63$ \\
    22.5\% & 0.98 & 1.62 & 2.97 & 15.65 & \textbf{15.00} & $-0.65$ \\
    50.6\% & 0.90 & 2.00 & 3.04 & 15.54 & \textbf{15.04} & $-0.50$ \\
    71.8\% & 0.63 & 2.34 & 3.33 & 15.43 & \textbf{14.95} & $-0.48$ \\
    80.1\% & 0.57 & 2.43 & 3.48 & 15.52 & \textbf{15.00} & $-0.52$ \\
    \midrule
    \multicolumn{7}{@{}l}{\emph{Wind flows, Wan2.2: test VRMSE}} \\
    0\% & 1 & 0.148 & 0.165 & 0.450 & \textbf{0.428} & $-4.9\%$ \\
    27.2\% & 0.97 & 0.153 & 0.163 & 0.482 & \textbf{0.449} & $-6.8\%$ \\
    51.9\% & 0.77 & 0.168 & 0.182 & 0.477 & \textbf{0.456} & $-4.4\%$ \\
    74.7\% & 0.65 & 0.216 & 0.223 & 0.493 & \textbf{0.475} & $-3.7\%$ \\
    \bottomrule
  \end{tabular}%
  }
\end{table}

\begin{table*}[t]
  \centering
  \small
  \caption{\textbf{Depth against published methods.} Zero-shot affine-invariant monocular depth: AbsRel ($\times100$) and pixels with $\delta_1<1.25$ (\%). Published numbers from~\cite{pavlovic2026marigoldv2}, Marigold: depth v1. Ours: the depth models of \cref{tab:final}, one sample per image, mean over four seeds.}
  \label{tab:suppl_final_depth}
  \resizebox{\textwidth}{!}{%
  \begin{tabular}{@{}lccccccccccccc@{}}
    \toprule
    & & \multicolumn{2}{c}{NYUv2} & \multicolumn{2}{c}{KITTI} & \multicolumn{2}{c}{ETH3D} & \multicolumn{2}{c}{ScanNet} & \multicolumn{2}{c}{DIODE} & \multicolumn{2}{c}{Mean} \\
    \cmidrule(lr){3-4}\cmidrule(lr){5-6}\cmidrule(lr){7-8}\cmidrule(lr){9-10}\cmidrule(lr){11-12}\cmidrule(l){13-14}
    Method & Images & AbsRel$\downarrow$ & $\delta_1{\uparrow}$ & AbsRel$\downarrow$ & $\delta_1{\uparrow}$ & AbsRel$\downarrow$ & $\delta_1{\uparrow}$ & AbsRel$\downarrow$ & $\delta_1{\uparrow}$ & AbsRel$\downarrow$ & $\delta_1{\uparrow}$ & AbsRel$\downarrow$ & $\delta_1{\uparrow}$ \\
    \midrule
    MoGe-2~\cite{wang2025moge2} & 8.9M & 3.1 & 98.4 & 6.9 & 89.9 & 5.0 & 96.2 & 3.1 & 98.3 & 9.6 & 84.5 & 5.54 & 93.46 \\
    Marigold~\cite{ke2024marigold} & 74K & 5.5 & 96.4 & 9.9 & 91.6 & 6.5 & 96.0 & 6.4 & 95.1 & 10.0 & 90.7 & 7.66 & 93.96 \\
    GeoWizard~\cite{fu2024geowizard} & 280K & 5.2 & 96.6 & 9.7 & 92.1 & 6.4 & 96.1 & 6.1 & 95.3 & 12.0 & 89.8 & 7.88 & 93.98 \\
    GenPercept~\cite{xu2025genpercept} & 90K & 5.2 & 96.6 & 9.4 & 92.3 & 6.6 & 95.7 & 5.6 & 96.5 & -- & -- & -- & -- \\
    Lotus~\cite{he2025lotus} & 59K & 5.4 & 96.8 & 8.5 & 92.2 & 5.9 & 97.0 & 5.9 & 95.7 & 9.8 & 92.4 & 7.10 & 94.82 \\
    Lotus-2~\cite{he2025lotus2} & 59K & 3.7 & 97.6 & 6.7 & 94.1 & 4.1 & 98.6 & 4.0 & 97.2 & 6.5 & 95.5 & 5.00 & 96.60 \\
    FE2E~\cite{wang2026fe2e} & 71K & 3.8 & 97.6 & 6.5 & 96.0 & 3.8 & 98.7 & 4.3 & 97.1 & 5.6 & 96.4 & 4.80 & 97.16 \\
    Marigold~V2~\cite{pavlovic2026marigoldv2} & 74K & 3.6 & 98.0 & 5.4 & 97.4 & 2.8 & 99.2 & 3.7 & 97.9 & 5.2 & 97.1 & 4.14 & 97.92 \\
    \midrule
    DPA, depth (ours) & 90K & 3.8 & 97.8 & 5.8 & 97.0 & 3.0 & 99.1 & 4.0 & 97.5 & 5.5 & 96.5 & 4.39 & 97.58 \\
    DPA, multi-modal (ours) & 60K & 4.0 & 97.8 & 6.4 & 96.1 & 3.8 & 98.6 & 4.1 & 97.3 & 6.2 & 96.1 & 4.89 & 97.16 \\
    \bottomrule
  \end{tabular}%
  }
\end{table*}

\begin{table*}[t]
  \centering
  \small
  \caption{\textbf{Surface normals against published methods.} Zero-shot mean angular error (degrees) and pixels within $11.25^\circ$ (\%). Published numbers from~\cite{pavlovic2026marigoldv2}. Ours: the normals models of \cref{tab:final}, one sample per image, mean over four seeds.}
  \label{tab:suppl_final_normals}
  \resizebox{\textwidth}{!}{%
  \begin{tabular}{@{}lcccccccccccc@{}}
    \toprule
    & & \multicolumn{2}{c}{NYUv2} & \multicolumn{2}{c}{ScanNet} & \multicolumn{2}{c}{iBims-1} & \multicolumn{2}{c}{Sintel} & \multicolumn{2}{c}{Mean} \\
    \cmidrule(lr){3-4}\cmidrule(lr){5-6}\cmidrule(lr){7-8}\cmidrule(lr){9-10}\cmidrule(l){11-12}
    Method & Images & Mean$\downarrow$ & $11.25^\circ{\uparrow}$ & Mean$\downarrow$ & $11.25^\circ{\uparrow}$ & Mean$\downarrow$ & $11.25^\circ{\uparrow}$ & Mean$\downarrow$ & $11.25^\circ{\uparrow}$ & Mean$\downarrow$ & $11.25^\circ{\uparrow}$ \\
    \midrule
    MoGe-2~\cite{wang2025moge2} & 8.9M & 14.7 & 62.3 & 12.8 & 68.4 & 14.7 & 70.4 & 29.3 & 24.8 & 17.88 & 56.48 \\
    DSINE~\cite{bae2024dsine} & 160K & 16.4 & 59.6 & 16.2 & 61.0 & 17.1 & 67.4 & 34.9 & 21.5 & 21.15 & 52.38 \\
    GeoWizard~\cite{fu2024geowizard} & 280K & 18.9 & 50.7 & 17.4 & 53.8 & 19.3 & 63.0 & 40.3 & 12.3 & 23.98 & 44.95 \\
    StableNormal~\cite{ye2024stablenormal} & 250K & 18.6 & 53.5 & 17.1 & 57.4 & 18.2 & 65.0 & 36.7 & 14.1 & 22.65 & 47.50 \\
    Diffusion-E2E-FT~\cite{garcia2025finetuning} & 74K & 16.5 & 60.4 & 14.7 & 66.1 & 16.1 & 69.7 & 33.5 & 22.3 & 20.20 & 54.63 \\
    GenPercept~\cite{xu2025genpercept} & 74K & 18.2 & 56.3 & 17.7 & 58.3 & 18.2 & 64.0 & 37.6 & 16.2 & 22.93 & 48.70 \\
    Lotus-G~\cite{he2025lotus} & 59K & 16.5 & 59.4 & 15.1 & 63.9 & 17.2 & 66.2 & 33.6 & 21.0 & 20.60 & 52.63 \\
    Lotus-D~\cite{he2025lotus} & 59K & 16.2 & 59.8 & 14.7 & 64.0 & 17.1 & 66.4 & 32.3 & 22.4 & 20.08 & 53.15 \\
    Marigold v1.1~\cite{ke2025marigold} & 77K & 16.1 & 60.5 & 14.5 & 66.1 & 16.3 & 68.5 & -- & -- & -- & -- \\
    Lotus-2~\cite{he2025lotus2} & 59K & 16.9 & 59.0 & 14.2 & 66.8 & 15.4 & 70.4 & 30.3 & 27.6 & 19.20 & 55.95 \\
    FE2E~\cite{wang2026fe2e} & 71K & 16.2 & 59.6 & 13.8 & 67.2 & 15.1 & 70.6 & 31.2 & 22.3 & 19.08 & 54.93 \\
    Marigold~V2~\cite{pavlovic2026marigoldv2} & 59K & 16.6 & 61.2 & 14.1 & 67.4 & 15.9 & 70.9 & 28.7 & 27.6 & 18.83 & 56.78 \\
    \midrule
    DPA, normals (ours) & 61K & 16.3 & 62.0 & 14.1 & 68.1 & 14.5 & 73.2 & 26.4 & 32.5 & 17.84 & 58.95 \\
    DPA, multi-modal (ours) & 60K & 16.4 & 61.7 & 14.0 & 68.3 & 14.6 & 72.7 & 25.4 & 34.7 & 17.61 & 59.34 \\
    \bottomrule
  \end{tabular}%
  }
\end{table*}

\section{VAE pruning}
\label{sec:suppl_prune}

\paragraph{Procedure.}
Following the depth pruning of diffusion transformers~\cite{fang2025tinyfusion}, we prune whole blocks, namely the residual blocks that keep the channel width and the attention layers of the mid blocks of the encoder and the decoder, and replace each removed block by the identity.
The blocks are removed greedily, each time the one whose removal increases the round-trip error on training data least per removed parameter, and the prefixes of this sequence are the pruned VAEs.
For the FLUX~3 VAE (84.0M parameters), the round-trip error is the masked L1 error on the normal maps of 16 Hypersim training images, and the prefixes we use remove 22.5\%, 50.6\%, 71.8\%, and 80.1\% of the parameters (4, 9, 16, and 19 blocks, L1 error 0.070, 0.142, 0.226, and 0.385 against 0.041 unpruned).
For the Wan2.2 VAE (704.7M parameters), it is the VRMSE of $(u,v)$ on 8 windows of 33 frames from 8 training simulations, and the prefixes remove 27.2\%, 51.9\%, and 74.7\% of the parameters (4, 11, and 16 blocks, VRMSE 0.51, 0.87, and 3.56 against 0.27).
Encoding and decoding times are measured for four $768\times768$ images, or one clip of 113 frames at $256\times256$, in bf16 on one RTX 6000 Ada.

\paragraph{Results.}
Each pruned VAE is fine-tuned without (control) and with DPA, followed by the DiT, as for the unpruned VAEs (\cref{tab:suppl_prune}).
On normals, the DPA VAEs with 71.8\% and 80.1\% of the parameters removed are as accurate as the unpruned DPA VAE ($14.95^\circ$ and $15.00^\circ$ vs.\ $14.98^\circ$) and encode and decode $1.6\times$ and $1.8\times$ faster.
The speed-up is smaller than the parameter reduction because most of the removed parameters belong to blocks at low resolution.
On wind flows, pruning raises the test VRMSE of the control from 0.450 to 0.482, 0.477, and 0.493, and DPA lowers it by 6.8\%, 4.4\%, and 3.7\% (4.9\% unpruned).
With 74.7\% of its parameters removed, the Wan2.2 VAE encodes and decodes a clip $1.5\times$ faster.

\section{Comparison with the state of the art}
\label{sec:suppl_sota}

\paragraph{Training.}
The depth and normals experiments of \crefrange{sec:exp_normals}{sec:exp_prune} train on Hypersim only and fine-tune the diffusion model for at most 5{,}000 steps to allow for running many experiments and comparisons at moderate cost.
For the comparison with the state of the art, the three models of \cref{tab:final} add Virtual KITTI~2, fine-tune their diffusion models for 20{,}000 steps, and end with the second stage of \cref{sec:exp_sota}, which optimizes the DiT and the VAE decoder together.
The depth model continues the diffusion model of a DPA model trained for 1{,}000 VAE and 2{,}200 diffusion steps on the depth data of Marigold~V2 (Hypersim and both Virtual KITTI~2 cameras at a ratio of 9:1, hence its 90K training frames) to 20{,}000 steps.
The normals model and its control add the 21{,}260 frames of one Virtual KITTI~2 camera (10\% of the batches) to the Hypersim data of \cref{sec:suppl_normals}, with 10{,}000 VAE and 20{,}000 diffusion steps, the schedule of Marigold~V2.
Before the second stage, the round-trip errors are $1.28^\circ$ for the control and $2.94^\circ$ for DPA, and with four samples averaged, DPA is $0.48^\circ$ more accurate on the indoor benchmarks ($14.83^\circ$ vs.\ $15.31^\circ$).
The multi-modal model continues the four-channel DPA model of \cref{tab:widen} for 3{,}000 VAE and about 17{,}700 diffusion steps on its 60K frames of Hypersim and Virtual KITTI~2, with Virtual KITTI~2 in 10\% to 25\% of the batches.
The VAE continuation uses learning rate $5\cdot10^{-5}$ and adds a gradient loss for depth (weight 0.5), the window-matched L1 loss~\cite{pavlovic2026marigoldv2} (0.2), and an angular loss for normals (0.1), with fixed weights outside GradNorm.

The second stage trains the LoRA of the diffusion model and the decoder through the two-step sampler at batch size 16 and learning rate $10^{-5}$, decayed to $10^{-6}$ over 3{,}000 steps (1{,}500 for the multi-modal model), with Virtual KITTI~2 in 25\% of the batches.
Its losses on the decoded maps are an L1 and a window-matched L1 loss per modality, a gradient loss for depth, and an angular loss for normals.
Unlike the diffusion loss, which acts on the latents, these losses act on the decoded maps that the benchmarks evaluate.
On depth, the second stage lowers the mean AbsRel ($\times100$) of the depth model from 4.57 to 4.39 (seed 0).
For the multi-modal model, we keep the checkpoint with the lowest depth error on 200 Hypersim validation frames, at step 300.

\paragraph{Recipe of Marigold~V2.}
The models of \cref{tab:final} use the diffusion loss and only the second stage of Marigold~V2.
On depth, we also tested its full recipe, one deterministic step from the latent of the input image with losses on the latent and on the decoded depth.
With 1{,}000 VAE and 2{,}200 diffusion steps on Hypersim and Virtual KITTI~2 (9:1), the control and DPA are equally accurate with the diffusion loss (AbsRel $\times100$ 5.16 and 5.25), whereas with this recipe DPA is less accurate (5.94 vs.\ 5.03, one seed each).
A likely reason is that a one-step regressor does not sample from the prior to which DPA anchors the latents, while the DPA VAE keeps its larger round-trip error.

\paragraph{Per-benchmark results.}
\Cref{tab:suppl_final_depth,tab:suppl_final_normals} list the errors of the models of \cref{tab:final} per benchmark, next to further published methods.
Among the methods trained on fewer than 5M images, which excludes MoGe-2, the depth model is second to Marigold~V2 on four of the five benchmarks and third on NYUv2.
On normals, FE2E is the most accurate of these methods on ScanNet ($13.8^\circ$ vs.\ $14.0^\circ$ for our multi-modal model), and our two models have the most pixels within $11.25^\circ$ on all four benchmarks.

\paragraph{Sintel.}
Sintel consists of frames of an animated film, unlike the indoor training data, and the models rank differently on it than on the indoor benchmarks (\cref{tab:suppl_sintel,tab:suppl_alternatives}).
For the models of \cref{tab:normals}, DPA is $1.99^\circ$ less accurate than the control on Sintel and $1.3^\circ$ less accurate than Marigold~V2 ($28.7^\circ$~\cite{pavlovic2026marigoldv2}), and the two are equally accurate on the mean of all four benchmarks ($+0.03^\circ$).
With VAEs pruned by 50.6\% and 71.8\%, DPA is $0.85^\circ$ and $2.35^\circ$ more accurate than the control on Sintel, and with 1\% to 25\% of the data $0.18^\circ$ to $0.36^\circ$ more accurate (means over three seeds).
Trained with Virtual KITTI~2, the models of \cref{tab:final} have the lowest Sintel errors of all methods in \cref{tab:suppl_final_normals} ($26.4^\circ$ and $25.4^\circ$).

\begin{table}[h]
  \centering
  \small
  \caption{\textbf{Normals on Sintel.} Mean angular error (degrees). Indoor: mean over iBims-1, NYUv2, and ScanNet. All: mean over the four benchmarks.}
  \label{tab:suppl_sintel}
  \resizebox{\columnwidth}{!}{%
  \begin{tabular}{@{}lccc@{}}
    \toprule
    VAE adaptation & Indoor$\downarrow$ & Sintel$\downarrow$ & All$\downarrow$ \\
    \midrule
    Control & 15.61 & 28.04 & 18.72 \\
    DPA & 14.98 & 30.03 & 18.75 \\
    \midrule
    Control / DPA, 22.5\% removed & 15.65 / 15.00 & 28.01 / 30.34 & 18.74 / 18.83 \\
    Control / DPA, 50.6\% removed & 15.54 / 15.04 & 29.47 / 28.62 & 19.03 / 18.44 \\
    Control / DPA, 71.8\% removed & 15.43 / 14.95 & 30.61 / 28.27 & 19.22 / 18.28 \\
    \bottomrule
  \end{tabular}%
  }
\end{table}

\begin{table}[t]
  \centering
  \small
  \caption{\textbf{MRI ablations.} Magnitude PSNR (dB) on the 96 test slices. Round trip: the VAE alone. (a), (c), and (d): 50 sampler steps.}
  \label{tab:suppl_mri}
  \begin{tabular}{@{}lcccccc@{}}
    \toprule
    \multicolumn{7}{@{}l}{(a) Anchor weight $\beta$} \\
    & 0 & 0.01 & 0.1 & 1 & & \\
    Test & 27.22 & 28.82 & \textbf{29.52} & 28.68 & & \\
    Round trip & 31.39 & 30.74 & 28.98 & 25.75 & & \\
    \midrule
    \multicolumn{7}{@{}l}{(b) Sampler steps (DPA with uniform levels)} \\
    & 2 & 5 & 10 & 25 & 50 & \\
    Control & 28.29 & 28.58 & \textbf{28.87} & 28.45 & 27.22 & \\
    DPA & 28.32 & 28.49 & 28.95 & 29.48 & \textbf{29.60} & \\
    \midrule
    \multicolumn{7}{@{}l}{(c) Training slices} \\
    & 8 & 32 & 304 & 608 & 1{,}216 & 4{,}256 \\
    Control & 28.07 & 27.75 & 27.74 & 27.65 & 27.51 & 27.22 \\
    DPA & 29.20 & 29.39 & 29.40 & 29.50 & 29.47 & 29.52 \\
    \midrule
    \multicolumn{7}{@{}l}{(d) VAE and diffusion fine-tuning steps, relative} \\
    & $\tfrac{1}{10}$ & $\tfrac{1}{4}$ & 1 & 2 & 4 & \\
    Control & 26.93 & 26.78 & 27.22 & 27.66 & 27.77 & \\
    DPA & 29.15 & 29.35 & 29.52 & 29.57 & 29.63 & \\
    \bottomrule
  \end{tabular}
\end{table}

\section{Accelerated MRI reconstruction}
\label{sec:suppl_mri}

\paragraph{Data.}
All MRI data are from the public, de-identified mridata.org repository~\cite{epperson2013mridata}.
Each slice is the coil-combined complex image at one readout position, divided by the 99.9th percentile of the magnitude of its volume, clipped to $[-1,1]$, and encoded as the three channels (real, imaginary, 0), so that all VAEs of \cref{tab:mri}, including the frozen one, take the same input.
Training uses positions 8--311 of the 14 training volumes (4{,}256 slices), and the test set the 32 central positions (144--175) of each test volume, each slice with its own mask seeded by its readout position.

\paragraph{Training.}
The control and DPA VAEs are fine-tuned for 1{,}000 steps at batch size 16 and learning rate $10^{-4}$ (50 warm-up steps, gradient clipping at 10, L1 loss, KL weight $10^{-6}$), DPA with $\beta=0.1$ and the noise floor.
FLUX~3 Image Klein Base is then fine-tuned with LoRA of rank 128 for 2{,}200 steps at batch size 16 on the latents of each VAE, including the frozen one, with a fixed prompt that describes a knee MRI slice.

\paragraph{Posterior sampling.}
At every step, the sampler decodes the clean estimate $\hat x$ of the diffusion model and makes it consistent with the measurements $y=M\odot\mathcal{F}(Sx)$, with the sensitivities $S$ of the 8 coils provided with the dataset, the 2D Fourier transform $\mathcal{F}$, and a variable-density Poisson-disc mask $M$ with a fully sampled $24\times24$ center.
Five conjugate-gradient steps from $\hat x$ on $\|M\odot\mathcal{F}(Sx)-y\|^2+\gamma\|x-\hat x\|^2$ give the consistent image, which is re-encoded and re-noised to the next level of the noise schedule of FLUX~3 with the noise predicted by the diffusion model instead of fresh noise.
After the last step, the measured $k$-space is restored exactly, and four samples are averaged.
The measurements are divided by 1.175 times the 99.9th percentile of the zero-filled magnitude, so that the scaling does not use the target.
$\gamma$ and the number of steps ($\{2,5,10,25,50\}$) are tuned on 18 validation slices (positions 40, 70, \ldots, 280 of the two validation volumes).

\paragraph{Evaluation and baselines.}
PSNR uses the 99.9th percentile of the target magnitude of the volume as peak.
The evaluation code of SMRD~\cite{ozturkler2023smrd} uses the maximum of each slice instead, which adds 5.65\,dB on average.
On that scale, our rerun of the Langevin sampler of~\cite{jalal2021robust}, with its code, knee settings, and score model trained on fastMRI brain~\cite{zbontar2018fastmri}, gives 31.45\,dB, matching the reported 31.4\,dB.
L1-wavelet uses SigPy~\cite{ong2019sigpy} with 100 iterations and a regularization weight of 0.01, tuned on validation over $[0.003,0.03]$ with the measured $k$-space restored (without the restoration, its best weight of 0.007 gives 29.02\,dB on test).

\paragraph{Pretrained diffusion model as prior.}
Posterior sampling with the original FLUX~3, without fine-tuning on MRI, reaches 25.32\,dB with the frozen VAE, 27.87\,dB with the decoder-only VAE, 27.46\,dB with the control, and 29.55\,dB with DPA.
With DPA, the original FLUX~3 thus matches the fine-tuned one (29.52\,dB) and is 2.09\,dB more accurate than with the control VAE, so the gain of DPA comes from aligning the latents with the pretrained prior rather than from fine-tuning the DiT.
Unlike surface normal estimation, where the DiT learns a conditional mapping between modalities, posterior sampling queries the unconditional prior to which DPA anchors the latents.

\paragraph{Ablations.}
\Cref{tab:suppl_mri} varies one setting at a time around \cref{tab:mri}, with 50 sampler steps unless stated otherwise.
Uniform noise levels instead of the mixture of \cref{sec:adjust} give 29.60\,dB ($+0.08$\,dB, better on 80 of 96 slices) and a 0.9\,dB better round trip.
A decoder-only stage after DPA (2{,}000 steps) without the noise floor improves the round trip by 1.4\,dB but lowers the reconstruction PSNR by 0.47\,dB, consistent with the role of the noise floor in \cref{sec:adjust}.

\section{Urban wind-flow simulation}
\label{sec:suppl_wind}

\paragraph{Setup.}
WinDiNet conditions LTX-Video on a first frame that encodes the building footprint and the inlet, and on the inlet speed and the domain size, and generates the velocity $(u,v)$ for 112 time steps at $256\times256$.
Each frame stores $(u,v)$, divided by 30\,m/s, and the building mask in its three channels, so the video VAEs need no changes to their input and output layers.
Our VAEs keep the physics loss on the decoded flow with which WinDiNet fine-tunes its decoder.
We use LTX-Video 2B (version 0.9.6-dev) and Wan2.2 (5B).

\paragraph{Training.}
The LTX-Video VAEs are fine-tuned on 5{,}010 training simulations for ten epochs at batch size 32 and learning rate $10^{-5}$ decayed to $10^{-6}$ (50 warm-up steps, gradient clipping at 10, KL weight $10^{-6}$), DPA with $\beta=10^{-2}$ and the noise levels and noise floor of the image experiments.
The diffusion model is then fully fine-tuned on all 10{,}000 training simulations at batch size 64 (WinDiNet: 16) and learning rate $10^{-5}$, on a cosine schedule over 10{,}000 steps stopped at step 3{,}900.
For Wan2.2, the VAEs are trained for five epochs on random windows of 33 frames, and the diffusion model on a cosine schedule over 3{,}600 steps, stopped after nine epochs (about 1{,}400 steps).
Rollouts use two sampler steps, and \cref{tab:wind} reports the mean over the last three checkpoints of each diffusion model, one epoch apart.

\paragraph{Inverse design.}
For inverse design, the layouts proposed by a surrogate must improve the simulated flow.
We test this in the morph mode of~\cite{perini2026windinet}, in which 15 buildings, each split into $2\times2$ blocks held together by a cohesion penalty, are moved by gradient descent through the surrogate to reduce the share of pedestrian-level wind speeds above 15\,m/s (danger) and 5\,m/s (discomfort) and below 1\,m/s (stagnation) in a target region, or to keep the speeds within a target band.
We optimize the five problems of~\cite{perini2026windinet} and 16 new ones (8 random layouts, 2 inlets each) through each LTX-Video surrogate of \cref{tab:wind}, with two sampler steps and 200 Adam steps (400 for two inlets), and simulate the initial and the optimized layouts with the CFD solver.
On the simulated flow, the layouts optimized through DPA lower the objective by 27.2\% on average, those of WinDiNet by 26.0\%, and those of the control by 25.7\%.

\section{Training cost}
\label{sec:suppl_cost}
On NVIDIA GH200 GPUs, the VAE stage takes 2.1 GPU-hours and the diffusion stage 12.3 GPU-hours for normals with FLUX~3, and 13.8 and 36.2 GPU-hours for wind flows with LTX-Video.
The anchor adds one forward pass of the frozen diffusion model per VAE step and no backward pass through it, which multiplies the VAE training time by 1.39 for FLUX~3 and by 1.04 for LTX-Video.

\end{document}